\documentclass[preprint,sort&compress,11pt,3p]{elsarticle}
\usepackage{moreverb}
\usepackage{mathtools}
\usepackage{amsmath}
\usepackage{amssymb}
\usepackage{algorithmic}
\usepackage{graphics}
\usepackage{graphicx}
\usepackage{subfigure}
\usepackage{caption}
\usepackage{enumitem}
\usepackage{extarrows}
\usepackage{color}
\usepackage{framed}
\usepackage{wrapfig}
\usepackage{bm}
\usepackage{mathrsfs}
\usepackage{mathabx}
\usepackage{multirow}
\usepackage{longtable}
\usepackage{hyperref}
\usepackage{paralist}
\usepackage{indentfirst}
\usepackage{relsize}
\usepackage{extarrows}
\usepackage{lineno,xcolor}
\usepackage{upgreek}
\usepackage{bm}
\usepackage{mwe}
\usepackage{xcolor}
\usepackage{booktabs}
\usepackage[flushleft]{threeparttable}
\usepackage[linesnumbered,lined,ruled]{algorithm2e}

\newcommand{\eref}[1]{(\ref{#1})}

\hypersetup{
bookmarks=true,
bookmarksopen=true,
bookmarksnumbered=true,
unicode=false,
pdftoolbar=true,
pdfmenubar=true,
pdffitwindow=false,
pdfstartview={FitH},
pdftitle={My title},
pdfauthor={Author},
pdfsubject={Subject},
pdfcreator={Creator},
pdfproducer={Producer},
pdfkeywords={keywords},
pdfnewwindow=true,
colorlinks=true,
linkcolor=blue,
citecolor=blue,
filecolor=magenta,
urlcolor=blue
}

\journal{Journal Name}

\begin{document}
	
\title{\Large PhyCRNet: Physics-informed Convolutional-Recurrent Network for Solving Spatiotemporal PDEs}

\author[NU1]{Pu Ren}
\ead{ren.pu@northeastern.edu}
\author[NU2]{Chengping Rao}
\author[NU2]{Yang Liu\corref{cor}}
\ead{yang1.liu@northeastern.edu}
\author[ND]{Jian-Xun Wang}
\author[NU1,MIT]{Hao Sun}
\ead{h.sun@northeastern.edu}
\cortext[cor]{Corresponding author. Tel: +1 617-373-3888}

\address[NU1]{Department of Civil and Environmental Engineering, Northeastern University, Boston, MA 02115, USA}
\address[NU2]{Department of Mechanical and Industrial Engineering, Northeastern University, Boston, MA 02115, USA}
\address[ND]{Department of Aerospace and Mechanical Engineering, University of Notre Dame, Notre Dame, IN 46556, USA}
\address[MIT]{Department of Civil and Environmental Engineering, MIT, Cambridge, MA 02139, USA}

\begin{abstract}
	\small
    Partial differential equations (PDEs) play a fundamental role in modeling and simulating problems across a wide range of disciplines. Recent advances in deep learning have shown the great potential of physics-informed neural networks (PINNs) to solve PDEs as a basis for data-driven modeling and inverse analysis. However, the majority of existing PINN methods, based on fully-connected NNs, pose intrinsic limitations to low-dimensional spatiotemporal parameterizations. Moreover, since the initial/boundary conditions (I/BCs) are softly imposed via penalty, the solution quality heavily relies on hyperparameter tuning. To this end, we propose the novel physics-informed convolutional-recurrent learning architectures (PhyCRNet and PhyCRNet-s) for solving PDEs without any labeled data. Specifically, an encoder-decoder convolutional long short-term memory network is proposed for low-dimensional spatial feature extraction and temporal evolution learning. The loss function is defined as the aggregated discretized PDE residuals, while the I/BCs are hard-encoded in the network to ensure forcible satisfaction (e.g., periodic boundary padding). The networks are further enhanced by autoregressive and residual connections that explicitly simulate time marching. The performance of our proposed methods has been assessed by solving three nonlinear PDEs (e.g., 2D Burgers' equations, the $\lambda$-$\omega$ and FitzHugh Nagumo reaction-diffusion equations), and compared against the start-of-the-art baseline algorithms. The numerical results demonstrate the superiority of our proposed methodology in the context of solution accuracy, extrapolability and generalizability.
\end{abstract}

\begin{keyword}
	\small
    Convolutional-recurrent learning \sep 
	Partial differential equations \sep 
	Encoder-decoder \sep 
	Physics-informed deep learning \sep 
	Residual connection \sep
	Hard-encoding of I/BCs 
\end{keyword}

\maketitle

\section{Introduction}\label{s:intro}
Complex spatiotemporal systems, modeled by PDEs, are ubiquitous in many disciplines such as applied mathematics, physics, biology, chemistry and engineering. Solving PDE systems has been a critical component in the community of scientific computing. Since the analytical solutions are inaccessible to most of the physical systems, numerical approaches have been extensively investigated and developed in recent decades, such as the finite difference/element/volume methods \cite{hughes2012finite} and isogeometric analysis (IGA) \cite{hughes2005isogeometric}. Although the classical numerical methods that approximate the exact solutions with basis functions and unknown parameters can achieve great accuracy for forward analysis, the computational demand remains a critical issue in applications of data assimilation and inverse problems, e.g., due to the requirement of repeated forward simulations. In the meanwhile, inevitable modeling errors and uncertainties are hard to adjust/mitigate in these intractable problems. 

Alternatively, recent developments in deep learning shed new lights on surrogate modeling of nonlinear systems for solving forward and inverse problems. The surrogate models leverage the powerful universal approximation capacity of deep neural networks (DNNs) \cite{hornik1989multilayer} which avoid repeated forward analyses and provide a promising direction for data assimilation and inverse problems. Herein, the core and fundamental challenge for NN-based approaches lies in how to effectively solve PDEs. Actually, dated back to last century, Lagaris \emph{et al.} \cite{lagaris1998artificial,lagaris2000neural} have already observed the similarities between spline-based numerical simulations and NN-based solutions for differential equations, and proposed the pioneering work on NN as the basic approximation element where the parameters are learnable by minimizing the physics-inspired error function. Over the past decade, thanks to the great advances in deep learning, many attempts have been devoted to this resurgent topic, which have led to a proliferation of studies in scientific machine learning \cite{huan2013simulation,han2018solving,zhang2018effect,raissi2019physics,bar2019learning,zhang2019efficient,zhu2018bayesian,brunton2020machine,gorodetsky2020mfnets,wang2021variational} with growing attention on modeling and simulation of PDE systems. Latest studies utilizing DNNs for modeling physical systems fall into two streams: continuous and discretized networks. The representative work of continuous learning is physics-informed neural networks (PINNs) \cite{raissi2019physics}, introduced for forward and inverse analysis of PDEs based on fully-connected DNNs. The general principle of PINNs inherits and matures the previous NN-based scheme \cite{lagaris1998artificial} with loss function consisting of PDE residuals, which facilitates DNNs training in ``small data" stage (e.g., small \cite{raissi2018deep,raissi2019deep,raissi2020hidden,qin2019data,samaniego2020energy,sun2020physics,rao2020physics-laminar} or zero labeled datasets \cite{sun2020surrogate,rao2020physics}). PINNs not only foster the current success in simulating various PDE systems (e.g., fluid dynamics \cite{yang2019predictive,jin2021nsfnets,mao2020physics,wessels2020neural}, solid mechanics \cite{rao2020physics,haghighat2021physics} and stochastic PDEs \cite{zhang2019quantifying,zhang2020learning}), but also show remarkable applications to a broad spectrum of other disciplines, including blood flows modeling \cite{kissas2020machine,cai2021artificial}, non-invasive inference \cite{shukla2020physics,yin2021non,lu2020extraction}, metamodeling of nonlinear structures \cite{zhang2020physics-cnn,zhang2020physics-lstm}, denoising \cite{he2020physics}, PDE discovery \cite{Chen2020DeepLO} and many others. Despite the great success and promise, there exists one central issue of scalability in the existing PINN framework: it is generally limited to low-dimensional parameterizations and less capable of handling PDE systems whose behavior has sharp gradients (e.g., propagating fronts of waves) or complex local morphology. The requirement of vast collocation points for PDE residuals may lead to huge computational cost and cause slow convergence in training \cite{gao2020phygeonet}. In addition, data-driven neural operators \cite{bhattacharya2020model,lu2021deepxde,lu2021learning,patel2021physics} also seek for the nonlinear mapping from parametric DNNs to the numerical solution in a continuous setting. The infinite-dimensional operator learning methods prevail over the tyranny of meshes/grids, but require a moderate amount of high-quality training data and tend to be computational intensively. 

A few very recent pilot studies show that physics-informed discrete learning schemes, e.g., convolutional neural networks (CNNs), possess better scalability and faster convergence \cite{zhu2018bayesian,winovich2019convpde,geneva2020modeling,gao2020phygeonet,ranade2021discretizationnet,bhatnagar2019prediction} for modeling PDE systems, thanks to their light-weight architecture and strength of efficient filtering over the computational domain. For the time-independent systems (e.g., steady-state PDEs), Zhu \emph{et al.} \cite{zhu2018bayesian,zhu2019physics} applied CNNs for surrogate modeling and uncertainty quantification (UQ) of PDE systems in rectangular reference domain. Furthermore, PhyGeoNet \cite{gao2020phygeonet} was proposed for geometry-adaptively solving steady-state PDEs via coordinate transformation between the physical and reference domains. On the other hand, for the time-dependent systems, the majority of the NN-based solutions still focus on data-driven approaches in the regular/rectangle grid \cite{sorteberg2018approximating,long2018pde,geneva2020transformers,li2020fourier} or the irregular mesh \cite{seo2019physics,trask2019gmls,li2020neural,belbute2020combining,pfaff2020learning}. Very few research (e.g., the AR-DenseED method in \cite{geneva2020modeling}) explores the possibility of using discrete learning to solve PDEs without any labeled data. Although the existing effort escapes the demanding requirement of high-quality training data, it has not shown satisfactory performance in error propagation \cite{geneva2020modeling}, due to the limitation of the basic autoregressive (AR) process. In general, relevant studies on scalable discretized learning architectures for solving spatiotemporal PDEs in ``small data '' regime remain limited in literature.

The specific objective of this paper is to propose a novel physics-informed convolutional-recurrent learning architecture (PhyCRNet) and its light-weight variant (PhyCRNet-s) for solving multi-dimensional spatiotemporal PDEs without any labeled data. We do not attempt to compare our proposed methods with classical numerical solvers, but instead provide a spatiotemporal deep learning perspective for surrogate modeling of complex PDEs, which can further serve as a basis approach for tackling challenges in data-enabled scientific computation such as inverse problems and data assimilation especially in sparse and noisy data regimes. The contributions of our paper are summarized as follows. First of all, the innovative PhyCRNet architecture combines the strengths of (1) an encoder-decoder convolutional long short-term memory network (ConvLSTM) \cite{shi2015convolutional} that extracts low-dimensional spatial features and learns their temporal evolution, (2) a global residual-connection that stringently maps the time-marching dynamics of the PDE solution, and (3) high-order finite-difference-based spatiotemporal filtering that accurately determines the essential PDE derivatives for constructing the residual loss function. Based on the fundamental neural components of PhyCRNet, we also propose PhyCRNet-s which periodically skips the encoder part in order to improve the computational efficiency. Secondly, hard-encoding of initial/boundary conditions (I/BCs) into the networks is implemented. The hard-imposed physical constraints drastically promote the solution accuracy on the boundaries. Finally, the numerical experiments ranging from nonlinear fluid dynamics to reaction-diffusion (RD) systems are performed to validate the proposed approaches. The numerical results show the superiority of PhyCRNet/PhyCRNet-s in the context of solution accuracy, extrapolability and generalizability in comparison with two baseline models. 

The rest of the paper is organized as follows, in addition to this Introduction section. Section \ref{s:prob} sets up the problem of solving PDE systems using DNNs. In Section \ref{s:method}, we elaborate the general principle and network architectures of PhyCRNet and PhyCRNet-s. In Section \ref{s:experiment}, we present the extensive numerical experiments and compare the performance between our networks and baseline methods. Section \ref{s:discussion} discusses the observations as well as the outlook of our future study. Section \ref{s:conclusion} concludes the entire paper.

\section{Problem Statement}\label{s:prob}
Herein, we consider the general form of a set of multi-dimensional ($n$), nonlinear, coupled PDE systems in parametric setting: 
\begin{equation}
    \label{eq:pde}
    \mathbf{u}_t + \mathcal{F}[\mathbf{u}, \mathbf{u}^2, \cdots, \nabla_{\mathbf{x}} \mathbf{u}, \nabla_{\mathbf{x}}^2 \mathbf{u}, \nabla_{\mathbf{x}} \mathbf{u} \cdot \mathbf{u}, \cdots; \boldsymbol{\lambda}] = \mathbf{0},
\end{equation}
where $\mathbf{u}(\mathbf{x},t) \in \mathbb{R}^{n}$ denotes the solution variable in the temporal domain $t \in [0,T]$ and the physical domain $\Omega$; $\mathbf{u}_t$ is the first-order time derivative term; $\nabla$ represents the gradient operator with respect to $\mathbf{x}$; $\mathcal{F}[\cdot]$ is the nonlinear functional parameterized by $\boldsymbol{\lambda}$. Additionally, the I/BCs have the form as $\mathcal{I}[\mathbf{u},\mathbf{u}_t;t=0,\mathbf{x} \in \Omega]=0$ and $\mathcal{B}[\mathbf{u},\nabla_{\mathbf{x}}\mathbf{u}, \cdots;\mathbf{x} \in \partial \Omega]=0$, where $\partial \Omega$ denotes the boundary of the spatial domain. 

Our general goal is to develop DNN-based methods for forward analysis of spatiotemporal PDE systems given in Eq. \eref{eq:pde}, which could serve as a basis for inverse problems when data is available (e.g., uncertainty quantification and data assimilation). More precisely, such networks will act as a new class of numerical solvers for various time-dependent PDEs given specific I/BCs. The entire training stage is unsupervised, where we do not require any labeled data and merely utilize the physical laws (e.g., PDEs and I/BCs) as constraints. Besides, to portray better local details in the solution, we discretize the physical domain and solve PDEs by employing (1) convolutional operators due to its faster convergence and better accuracy compared with fully-connected neural networks according to previous studies \cite{zhu2019physics,geneva2020modeling,gao2020phygeonet} and (2) recurrent units for controlling error propagation. In this work, we mainly focus on regular (e.g., rectangular) physical domains, where both the spatial and temporal domains are discretized uniformly and convolutional filtering can be applied in nature. Our objective is to point-wisely approximate the discrete solution field $\mathbf{u}^{\theta}=\mathbf{u}(\mathbf{x},t;\boldsymbol{\theta})$, that satisfies Eq. \eref{eq:pde} for specifically given I/BCs, where $\boldsymbol{\theta}$ denotes the network trainable parameters. Furthermore, we view the network training as an optimization process by minimizing the loss function composed of the discrete PDE residuals subjected to I/BCs. The details are described in Section \ref{s:method}.

\section{Methodology}\label{s:method}
In this section, two network architectures are proposed for solving spatiotemporal PDE systems. Based on the solid mathematical foundation in sparse representation for PDEs \cite{schaeffer2013sparse}, we attempt to extract the low-dimensional spatial features from dynamics and learn the temporal evolution on the compressed information through an encoder-decoder convolutional-recurrent scheme. The light-weight networks are built with the input of previous state variable and the output of quantities of interest (next state variable). Previously, Geneva and Zabaras \cite{geneva2020modeling} claimed that recurrent neural networks (e.g., long short-term memory) are powerful tools for predicting time-series whereas trapped in the difficulty of training for solving PDEs. However, we show that, based on our network setting, such an issue can be mitigated.

\subsection{ConvLSTM}\label{s:convlstm}
ConvLSTM is a spatiotemporal sequence-to-sequence learning framework extended from long short-term memory (LSTM) \cite{hochreiter1997long} and its variant LSTM encoder-decoder forecasting architecture \cite{graves2013generating,sutskever2014sequence}, which hold the strength of modeling the long-period dependencies that evolve in time. This great success comes from the distinctive and innovative memory cell and gated scheme of LSTM. Essentially, the memory cell is updated through the input and state information being accessed, accumulated and removed due to the delicate design of controlling gates. Based on such setup, the notorious gradient vanishing problem of vanilla recurrent neural networks (RNNs) has been relieved. It is also worthwhile to mention that a typical RNN model can be seen as a variant to a state-space model with nonlinear activation functions incorporated  \cite{jordan1997serial,DBLP:conf/nips/ChenRBD18}. Herein, LSTM, a special class of RNNs, acts as an implicit numerical scheme for solving time-dependent differential equations.   

\begin{figure}[t!]
	\centering
	    \subfigure[The single ConvLSTM cell at time $t$.]{\includegraphics[width=0.7\linewidth]{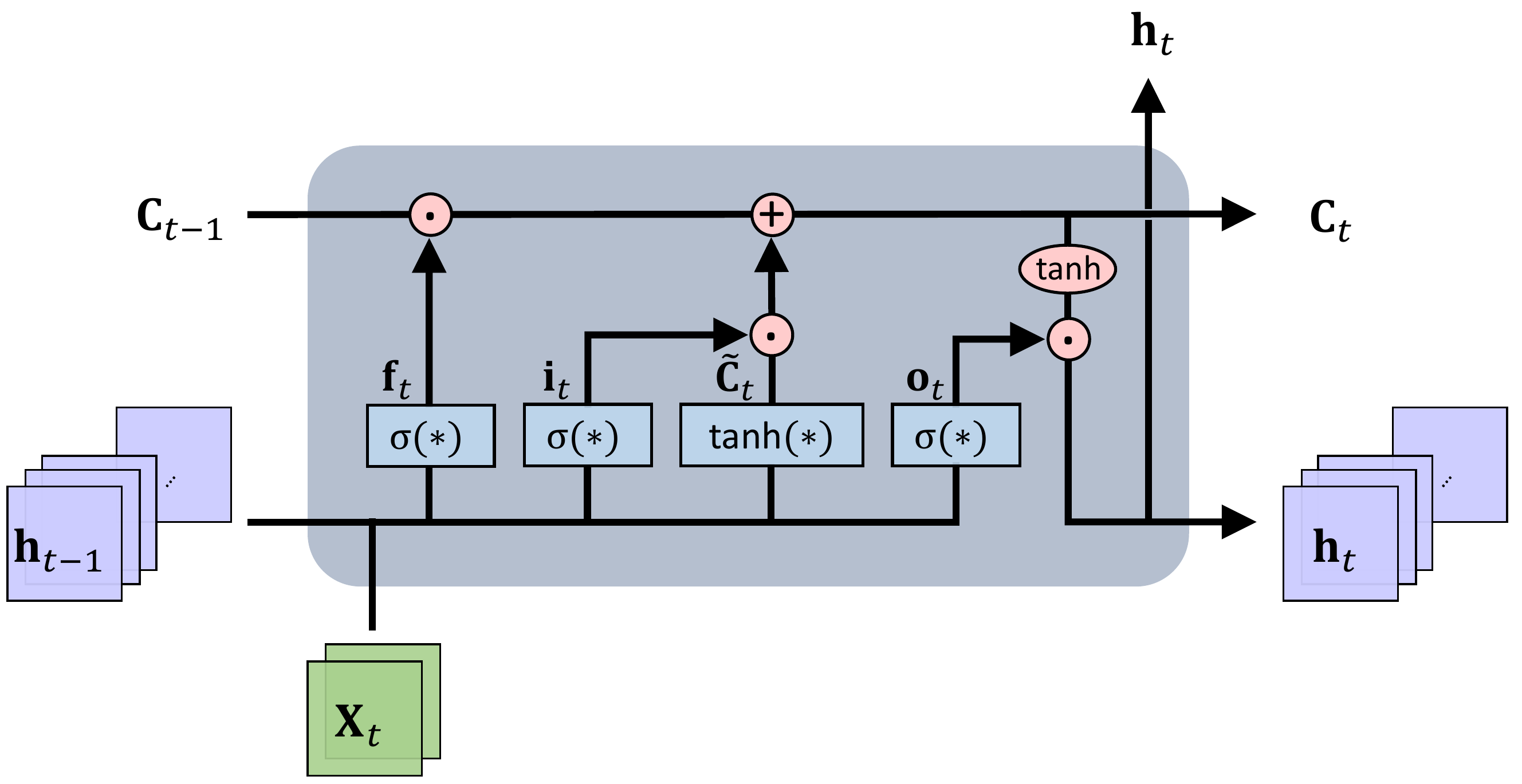}
	        \label{fig:convlstm}} 
	    \hspace{2em}
	    \subfigure[The pixel shuffle layer.]{\includegraphics[width=0.53\linewidth]{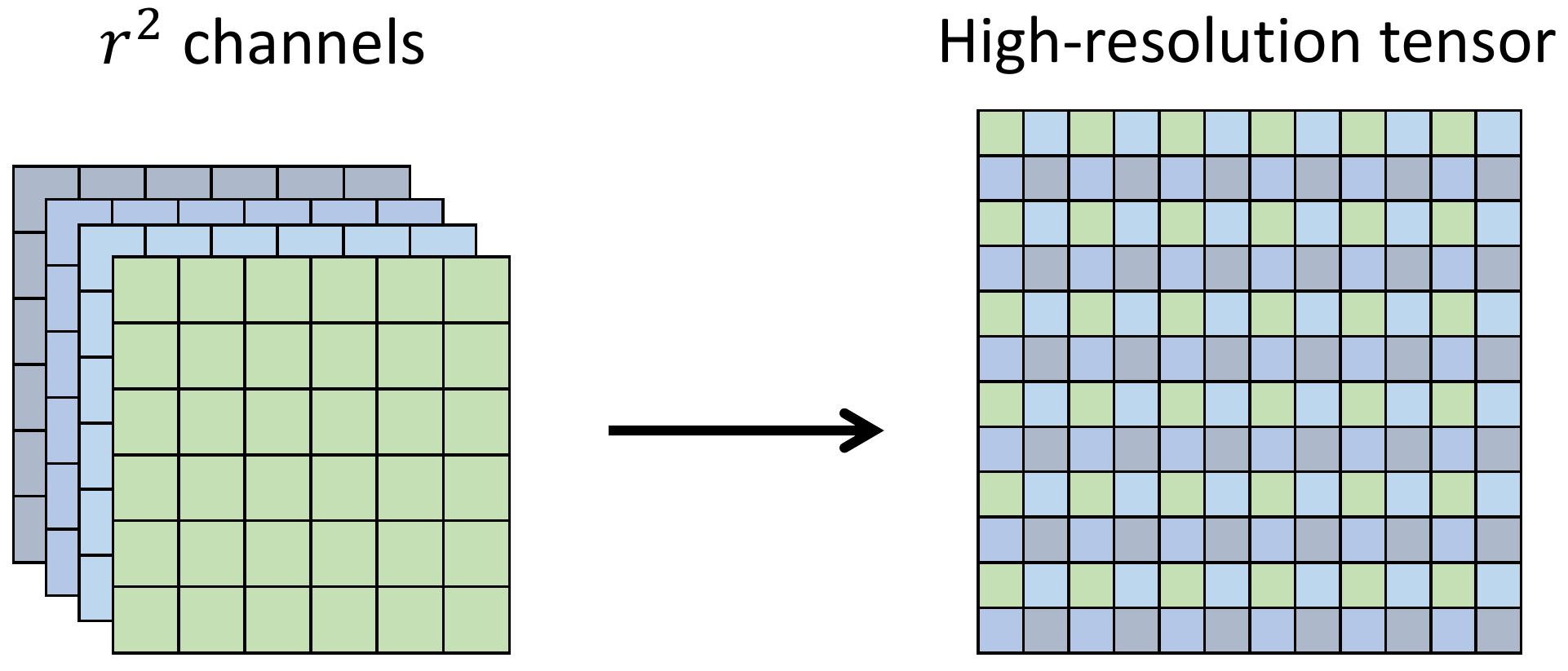}
	        \label{fig:pixelshuffle}} 
	\caption{The graphic illustration of network components.}
	\label{fig:component}
\end{figure}

The fundamental of ConvLSTM is to inherit the basic construction of LSTM (i.e., cells and gates) for controlling the information flow, and to modify the fully-connected NNs (FC-NNs) in gated operations to CNNs considering their better representational capability of spatial connections \cite{shi2015convolutional}. The graphic demonstration is presented in Figure \ref{fig:convlstm}. Let $\mathbf{X}_t$ denote the input tensor, and $\{\mathbf{h}_t,\mathbf{C}_t\}$ the hidden state and cell state to be updated at time $t$ respectively. Moreover, the ConvLSTM cell consists of four gate variables for input-to-state and state-to-state transitions, including a forget gate $\mathbf{f}_t$, an input gate $\mathbf{i}_t$, an internal cell $\widetilde{\mathbf{C}}_t$ and an output gate $\mathbf{o}_t$. In specific, due to the sigmoid activation function $\sigma(\cdot)$ mapping outputs to values between 0 and 1, the forget gate layer adaptively clears the memory information in the cell state $\mathbf{C}_{t-1}$. The memory stored in cell state originates from the cooperation between the input gate layer and the internal cell state, where the internal cell state is a new cell candidate created from hyperbolic tangent activation layer (i.e., tanh($\cdot$)) and the input gate layer decides the information propagating into the cell state. Lastly, the output gate layer filters and regulates the cell state for the final output variable/hidden state. The mathematical formulations of updating ConvLSTM cells are expressed as:
\begin{equation}
    \begin{split}
        \label{eq:convlstm}
        \mathbf{i}_t &= \sigma (\mathbf{W}_{i} * [\mathbf{X}_t,\mathbf{h}_{t-1}] + \mathbf{b}_i), \quad \quad
        \mathbf{f}_t = \sigma (\mathbf{W}_{f} * [\mathbf{X}_t,\mathbf{h}_{t-1}] + \mathbf{b}_f), \\ 
        \widetilde{\mathbf{C}}_{t-1} &= \text{tanh}(\mathbf{W}_{c} * [\mathbf{X}_t,\mathbf{h}_{t-1}] + \mathbf{b}_c), \quad  
        \mathbf{C}_t = \mathbf{f}_t \odot \mathbf{C}_{t-1} + \mathbf{i}_t \odot \widetilde{\mathbf{C}}_{t-1}, \\
        \mathbf{o}_t &= \sigma (\mathbf{W}_{o} * [\mathbf{X}_t,\mathbf{h}_{t-1}] + \mathbf{b}_o), \quad \quad
        \mathbf{h}_t = \mathbf{o}_t \odot \text{tanh}(\mathbf{C}_t), \\ 
    \end{split}
\end{equation}
where $*$ is the convolutional operation and $\odot$ denotes the Hadamard product; $\{\mathbf{W}_i,\mathbf{W}_f,\mathbf{W}_c,\mathbf{W}_o\}$ are the weight parameters for the corresponding filters while $\{\mathbf{b}_i,\mathbf{b}_f,\mathbf{b}_c,\mathbf{b}_o\}$ represent bias vectors.

\subsection{Pixel Shuffle}\label{s:pixelshuffle}
Widely adopted in super-resolution tasks, pixel shuffle is an efficient operation to conduct the sub-pixel convolutions \cite{shi2016real} in purpose of upscaling the low-resolution (LR) feature maps into the high-resolution (HR) outputs. The basic principle of pixel shuffle is described in Figure \ref{fig:pixelshuffle}. Let us consider a LR feature tensor of shape $(C\times r^2,H,W)$, where $C$ denotes the number of channels, $\{H,W\}$ refer to the height and width respectively, and $r$ is the upscaling factor. Straightforwardly, it realigns the elements in LR tensor to a HR tensor of shape $(C,H\times r,W\times r)$. 

Through a simple and fast operation, pixel shuffle maintains the satisfactory reconstruction accuracy in image and video super-resolution tasks without high computational and memory cost. There are two tricks contributing to the efficiency. Firstly, we can implement the sub-pixel convolution in the last layer for the spatial upscaling. It has lower computational complexity compared with other classical upscaling methods (e.g., deconvolution \cite{shan2008fast}) which always need more layers to reach expected resolution. Secondly, before the upsampling layer, all the feature extraction layers are based on the LR tensors where smaller filter can be employed. Beyond that, another advantage of pixel shuffle is that it introduces fewer checkerboard artifacts compared with deconvolution \cite{odena2016deconvolution}. Hence, we consider pixel shuffle operation as a preferable upsampling strategy in the network conception described in Subsection \ref{s:phycrnet}.

\subsection{Network architecture: PhyCRNet}\label{s:phycrnet}
\begin{figure}[b!]
    \centering
    \includegraphics[width=0.8\linewidth]{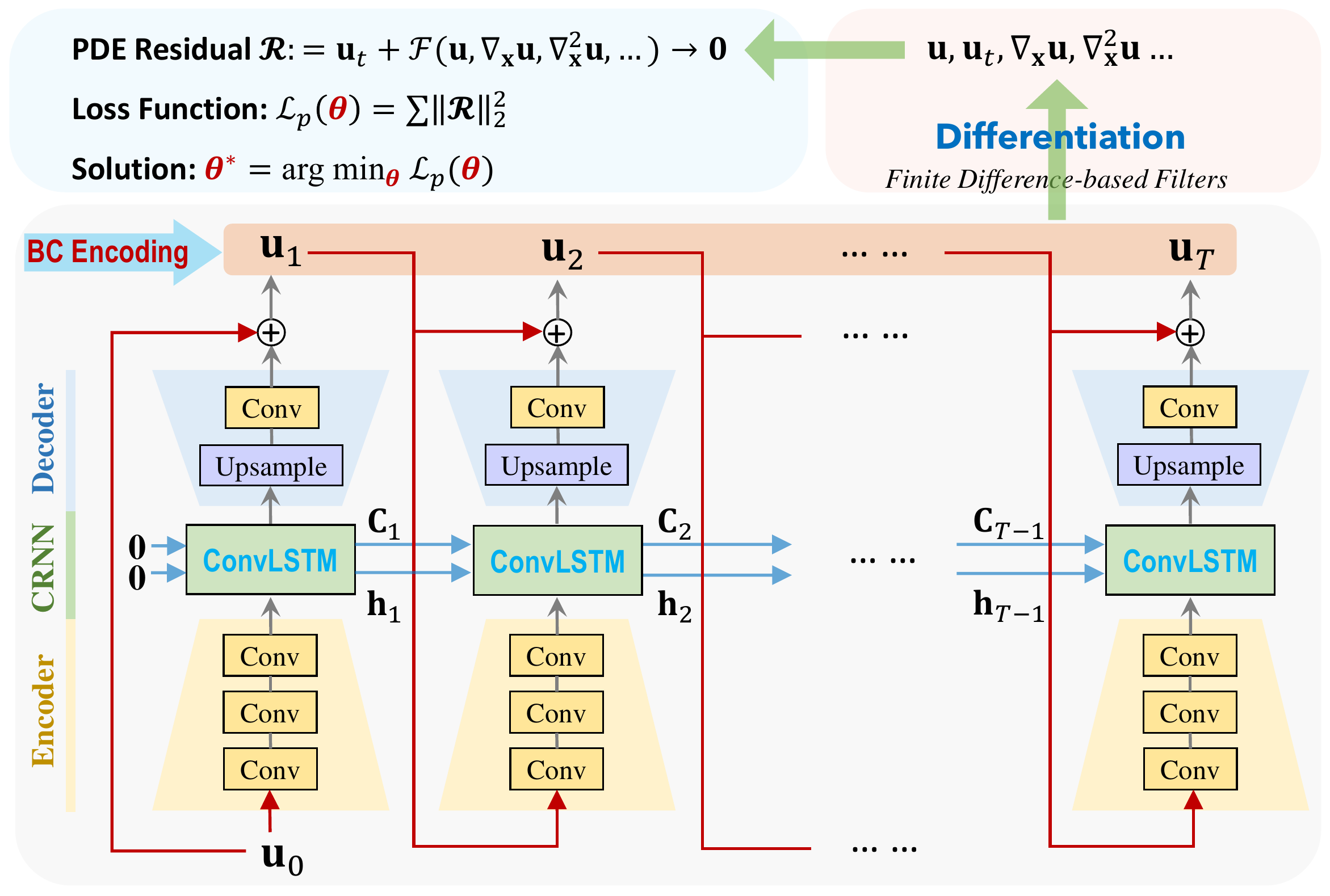}
    \caption{The network architecture of PhyCRNet. The variables {\bf C} and {\bf h} are cell state and hidden state of ConvLSTM. $\mathbf{u}_0$ is the known state variable (i.e., IC). ``BC Encoding'' means hard enforcement of BCs on the learned output variables after each training epoch, which serves for differentiation. Besides, $\boldsymbol{\theta}$ denotes the unknown trainable parameters in PhyCRNet.}
    \label{fig:phycrnet}
\end{figure}

In this part, we present the architecture of PhyCRNet, comprised of an encoder-decoder module, residual connection, autoregressive (AR) process and filtering-based differentiation. The framework is shown in Figure \ref{fig:phycrnet}. The encoder includes three convolutional layers for learning low-dimensional latent features from the input state variable $\mathbf{u}_i$ ($i=0,1,...,T-1$), where $T$ denotes the total number of time steps. We apply ReLU as the activation function for the convolutional layers. Then ConvLSTM layers act as the temporal propagator on the low-resolution latent features with the initial hidden/cell states starting at rest (e.g., $\mathbf{C}_0=\mathbf{0}$ and $\mathbf{h}_0=\mathbf{0}$). Modeling the essential dynamics on low-dimensional variables is capable of capturing the temporal dependency accurately and, meanwhile, helps alleviate the memory burden. Another strength of using LSTM comes from the hyperbolic tangent function for the output state, which holds smooth gradient curve and also pushes the values to be between $-1$ and $1$. Thus, after establishing the convolutional-recurrent scheme in the center, we directly reconstruct the low-resolution latent space to the high-resolution quantities based on an upsampling operation. In particular, the sub-pixel convolution layer (i.e., pixel shuffle) is applied due to its preferable efficiency and reconstruction accuracy with fewer artifacts compared with deconvolution. In the end, we add another convolutional layer for scaling the bounded output back to the original magnitude of interest. There is no activation function behind this scaling layer. Besides, it is worthwhile to mention that we do not consider batch normalization \cite{ioffe2015batch} in PhyCRNet in view of the limited amount of input variables and its deficiency to super-resolution \cite{yu2018wide}. As a substitute, we train the network with weight normalization \cite{salimans2016weight} for training acceleration and better convergence \cite{yu2018wide}.

Inspired by the forward Euler scheme, we append a global residual connection between the input state variable $\mathbf{u}_{i}$ and the output variable $\mathbf{u}_{i+1}$. The learning process at time instant $t_{i}$ is formulated as $\mathbf{u}_{i+1} = \mathbf{u}_{i} + \delta t \cdot \mathcal{NN}[\mathbf{u}_{i};\boldsymbol{\theta}]$, where $\mathcal{NN}[\cdot]$ denotes the trained network operator and $\delta t$ is the time interval. The output state variable $\mathbf{u}_{i+1}$ at time instant $t_{i}$ switches into the input variable at $t_{i+1}$. Actually, such input-output flow can be seen as a simple AR process (i.e., AR(1)). 

Hence, not only do we have temporal evolution in the central latent representation, but also build the propagation on the input and output at each time instant. Moreover, the introduction of ConvLSTM can also help moderate the rigorous time-stepping issues, where larger time intervals may be adopted compared with traditional numerical methods (see Section \ref{s:experiment}). Here, $\mathbf{u}_0$ is the given IC, and $\mathbf{u}_1, \mathbf{u}_2, \cdots, \mathbf{u}_{T}$ are the discrete solution variables to be predicted. Next, the remaining challenge is how to calculate the derivative terms. We apply the gradient-free convolutional filters to represent the discrete numerical differentiation in order to approximate the derivative terms of interest \cite{zhu2018bayesian,gao2020phygeonet}. For example, the finite-difference-based filters we considered in this paper are the second-order (see Eq. \eref{eq:diff_tem_filer}) and fourth-order central difference schemes (see Eq. \eref{eq:diff_spatial_filter}) to compute temporal and spatial derivatives, respectively, given by
\begin{equation}
    \label{eq:diff_tem_filer}
    K_t = [-1,0,1] \times \frac{1}{2\delta t},
\end{equation}
\begin{equation}
    \label{eq:diff_spatial_filter}
    K_s = {
    \begin{bmatrix} 
        0 & 0 & -1 & 0 & 0 \\
        0 & 0 & 16 & 0 & 0 \\
        -1 & 16 & -60 & 16 & -1 \\
        0 & 0 & 16 & 0 & 0 \\  
        0 & 0 & -1 & 0 & 0 
    \end{bmatrix}} \times \frac{1}{12(\delta x)^2}.
\end{equation}
where $\delta t$ and $\delta x$ denote the time spacing and spatial mesh size. Note that the derivatives on the boundaries cannot be computed directly due to the intrinsic deficiency of finite difference methods. The risk of losing differential information on the boundaries can be mitigated by designated padding mechanism (e.g., periodic padding) introduced in Subsection \ref{s:ibc}.

\subsection{Efficient network architecture: PhyCRNet-s}\label{s:phycrnet-s}
\begin{figure}[t!]
    \centering
    \includegraphics[width=0.97\linewidth]{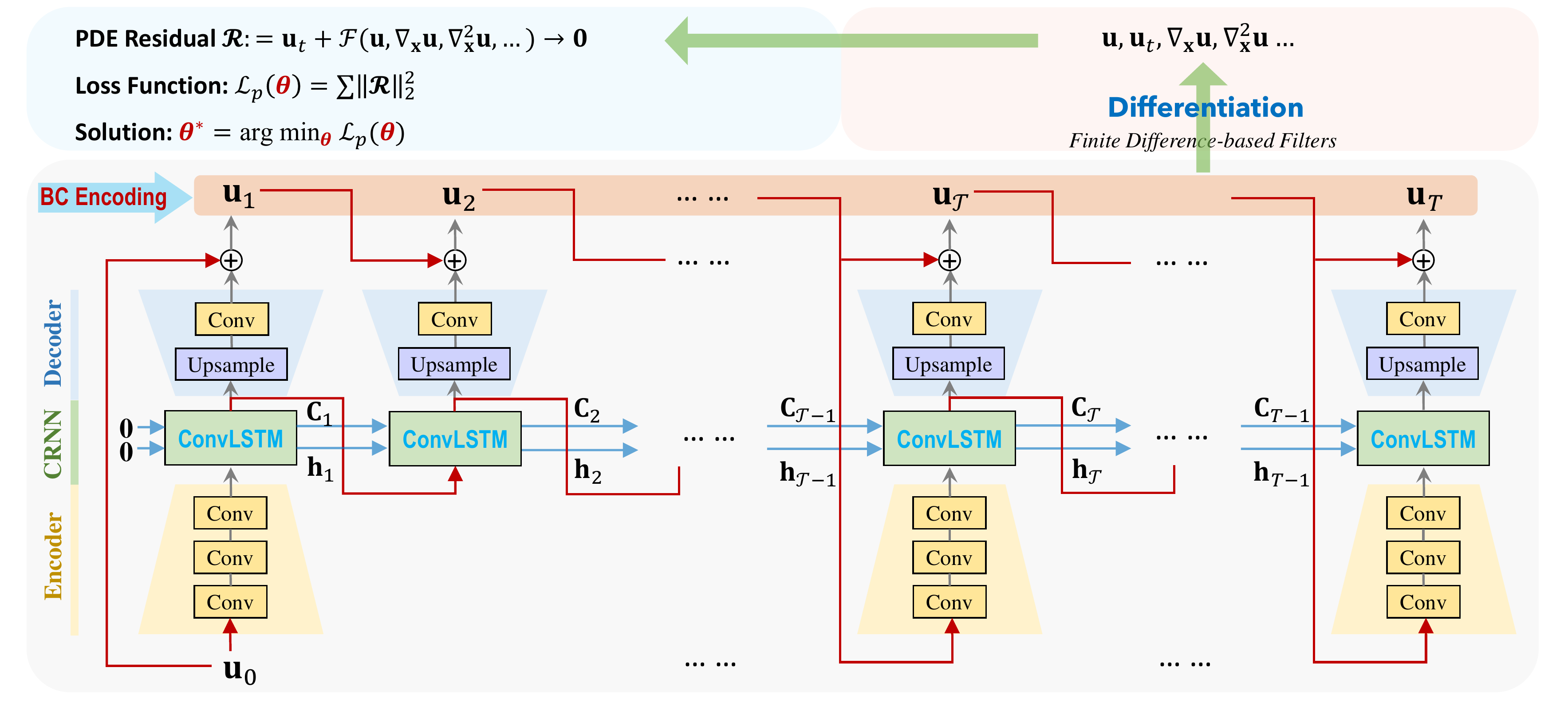}
    \caption{The network architecture of PhyCRNet-s. $\mathcal{T}$ is the cycle index of skipping-encoder.}
    \label{fig:phycrnet-s}
\end{figure}

To further improve the computational efficiency, we propose the second network architecture: PhyCRNet-s. We keep the majority of neural components in PhyCRNet, except skipping the encoder part periodically and setting a different input flow. Here we introduce a cycle index of skipping-encoder $\mathcal{T}$ ($\mathcal{T} \leq T$) for removing the redundant computation of the encoder. For instance, we have the HR input at time instant $0$ and $\mathcal{T}-1$ (i.e., $\mathbf{u}_0$ and $\mathbf{u}_{\mathcal{T}-1}$) where the encoder part of PhyCRNet is implemented, whereas the LR output feature from ConvLSTM is directly transmitted to the next step as LR input during the time period $[1,\mathcal{T}-2]$. The graphic illustration of skipping-encoder scheme is shown in Figure \ref{fig:phycrnet-s}. Since the encoder-decoder component shares the same learnable parameters and is not involved in temporal evolution, it is derived that the sparse residual dynamics has been learned in ConvLSTM. Namely, at time instant $t_i$, ConvLSTM works as a sparse dynamics propagator between the LR input feature $\mathbf{X}_i$ and the LR output variable $\mathbf{X}_{i+1}$. Then $\mathbf{X}_{i+1}$ can be naturally regarded as the LR input at time instant $t_{i+1}$. Similarly, such a construction also implicates the forward Euler scheme on sparse dynamics: $\mathbf{X}_{i+1}\approx \mathbf{X}_{i}+\delta t \cdot \mathcal{NN}^{c}[\mathbf{X}_i;\boldsymbol{\theta}^c]$ where $\mathcal{NN}^{c}[\cdot]$, denotes the ConvLSTM network and $\boldsymbol{\theta}^c$ is its corresponding trainable parameters. 

Although it brings a more light-weight architecture, this skipping-encoder scheme may cause approximation error in the temporal propagation. To seek for a tradeoff between accuracy and efficiency, we tend to set $\mathcal{T}$ as a relatively small value in case of error accumulation. The selection of $\mathcal{T}$ is empirically discussed in Subsection \ref{s:net_comp}. 

\begin{figure}[t!]
    \centering
    \includegraphics[width=0.8\linewidth]{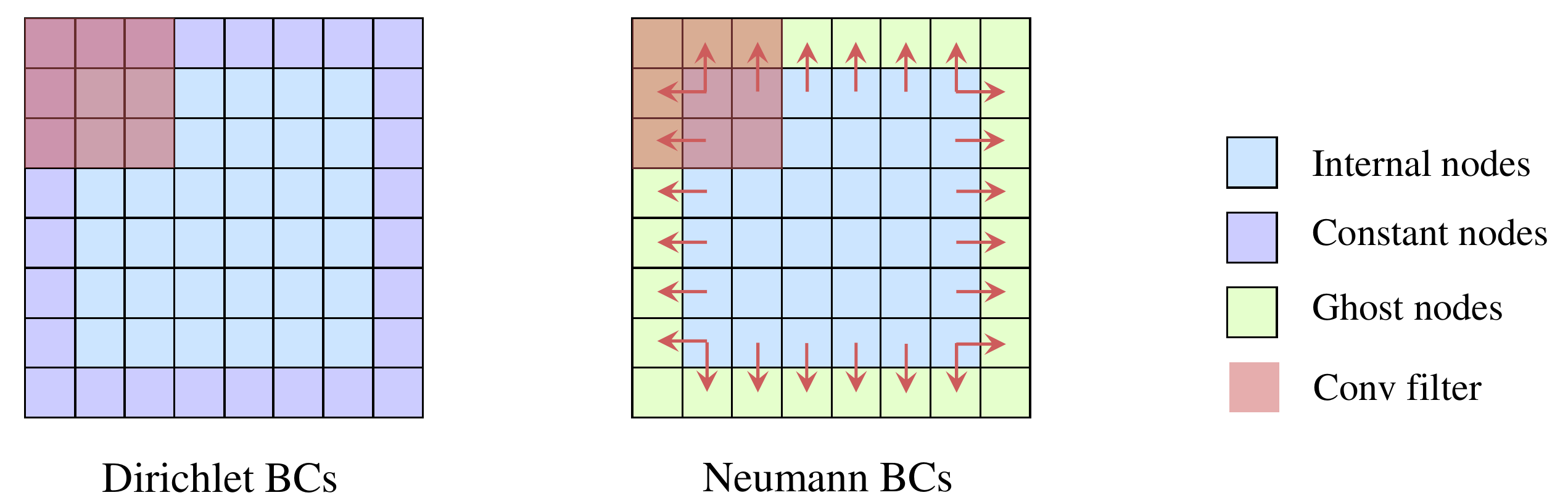}
    \caption{A graphic demonstration of hard imposition of BCs into the network. For the Dirichlet BCs (left figure), the constant nodes are padded on the boundaries. For the Neumann BCs (right figure), the ghost nodes are padded with values derived from internal field. Moreover, we implement the hard enforcement of BCs before the convolution operations both in the learning process and the differentiation.}
    \label{fig:BCs}
\end{figure}

\subsection{Hard imposition of I/BCs}\label{s:ibc}
The motivation of hard imposition of I/BCs is to prompt a well-posed optimization problem when training the network, which contributes to the improvement of solution accuracy and the facilitation of convergence \cite{gao2020phygeonet,rao2021embedding,rao2021hard}. Furthermore, the general philosophy behind it is to strictly integrate the known information of I/BCs into the network. Note that the IC can be easily imposed through the residual connection shown in Figure \ref{fig:phycrnet} and \ref{fig:phycrnet-s}. Since the solution field is uniformly discretized, it is suitable to apply padding for message passing pixel-wisely. For Dirichlet BCs, the known constants on the boundaries can be rigorously incorporated into the state variables via time-invariant padding operation. For Neumann BCs, ghost elements \cite{hughes2012finite} are necessary for forcible satisfaction beyond the boundaries, whose values can be approximately inferred with finite difference. Herein, the relationship between ghost nodes and internal nodes is time-invariant, but the values for padding change during the training process. The idea of hard imposition of I/BCs is illustrated in Figure \ref{fig:BCs}.

Specifically, we consider solving PDEs with periodic Dirichlet BCs in this work. Hence, the periodic padding (also known as circular padding) is introduced here for handling the constant copy of boundary values. The boundary padding operation aims to ensure that the minimization of PDE residuals proceeds in the entire solution domain without sacrificing any boundary nodes. Here we extend the periodic padding to all the feature maps produced by convolutional operations including those in ConvLSTM. We find it helps boost the solution accuracy on the boundaries compared with zero-padding. In addition, the periodic padding will be implemented twice on the boundaries when applying the fourth-order finite-difference-based filtering for differentiation.

\subsection{Physics-informed loss function}\label{s:loss_func}
Thanks to the hard enforcement of I/BCs, we only need to construct the loss function based on the governing PDEs. Firstly, we define the PDE residual $\boldsymbol{\mathcal{R}}(\mathbf{x},t; \boldsymbol{\theta})$ given by 
\begin{equation}
    \label{eq:pde_residual}
    \boldsymbol{\mathcal{R}}(\mathbf{x},t; \boldsymbol{\theta}):=\mathbf{u}_t^\theta + \mathcal{F}[\mathbf{u}^\theta, \nabla_{\mathbf{x}} \mathbf{u}^\theta, \nabla_{\mathbf{x}}^2 \mathbf{u}^\theta, \cdots; \boldsymbol{\lambda}],
\end{equation}
which is exactly the left-hand-side of Eq. \eref{eq:pde} with quantities of interest learned by the network. Furthermore, the shared network parameters $\boldsymbol{\theta}$ can be trained by minimizing the loss function $\mathcal{L}(\boldsymbol{\theta})$, which is equivalent to the sum of squares of the PDE residuals over the spatiotemporal discretization. Taking a two-dimensional PDE system as an example, $\mathcal{L}(\boldsymbol{\theta})$ can be expressed as 
\begin{equation}
    \label{eq:loss}
    \mathcal{L}(\boldsymbol{\theta}) = \sum_{i=1}^n\sum_{j=1}^m\sum_{k=1}^{T}\|\boldsymbol{\mathcal{R}}(x_i,y_j,t_k; \boldsymbol{\theta})\|_2^2,
\end{equation}
where $n$ and $m$ denote the height and width in the spatial grid, $T$ is the total number of time steps, and $\|\cdot\|_2$ denotes $\ell_2$ norm.

\section{Numerical experiments} \label{s:experiment}
In this section, we evaluate the performance of our proposed methods on three nonlinear PDE systems, and compare them with two baseline algorithms: the vanilla PINN approach \cite{raissi2019physics} and the AR-DenseED model \cite{geneva2020modeling}. These numerical experiments cover the Burgers' equations and two RD systems (i.e., $\lambda$-$\omega$ and FitzHugh-Nagumo equations) in 2D domains with periodic BCs. The specific experiments include three groups: (1) comparing the solution accuracy and extrapolability of PhyCRNet with baseline algorithms; (2) testing the generalization ability of PhyCRNet to different ICs; (3) comparing the performance between PhyCRNet and PhyCRNet-s. For all of these experiments, the spatial resolutions of input and output state variables are set as $128\times 128$. All the numerical implementations in this paper are coded in Pytorch \cite{paszke2017automatic} and performed on a NVIDIA Tesla V100 GPU card (32G) in a standard workstation\footnote{Source codes/datasets are available on GitHub at \url{https://github.com/isds-neu/PhyCRNet} upon  publication.}. 

\subsection{Network setup}\label{s:setup}
Herein, we consider the same network setting, of PhyCRNet and PhyCRNet-s, for all PDE cases. The encoder consists of three convolutional layers with 8, 32, 128 units respectively, using $4\times4$ kernels and a stride of 2. The periodic padding is applied to all convolutional operations. Afterwards, one ConvLSTM layer is merged on the latent space with cell/hidden states of 128 nodes. The convolutional operations in ConvLSTM have $3\times3$ kernels and a stride of 1. For the scaling layer before the output, we use a larger kernel ($5\times5$) and the same stride in consideration of filtering on high-resolution spatial features. The upsampling factor for pixel-shuffle is 8. The networks are trained by the stochastic gradient descent Adam optimizer \cite{kingma2014adam}. We train PhyCRNet for 10,000 epochs based on a shorter period of evolution, e.g., $T/3\sim2T/3$.

\subsubsection{Baseline model setup}
The training implementations of baselines are based on the reasonable neural architectures and training efforts similar to our proposed methods for fair comparison. Besides, we apply the same network parameters of the baselines to solve all PDE systems. We mainly compare the solution snapshots between PhyCRNet and PINN as a typical case of discrete and continuous learning, and compare the error propagation among PhyCRNet, PINN and AR-DenseED.  

The PINN model has 4 fully-connected layers, each with 80 nodes. A total of $1.62\times10^{5}$ collocation points are used to evaluate the total loss function (e.g., PDE residual loss + I/BC loss), trained by $1\times10^{4}$ epochs using Adam followed by up to $1\times10^{5}$ epochs with the L-BFGS optimizer \cite{liu1989limited}. For the AR-DenseED method which incorporates the AR process into the DenseNet architecture \cite{huang2017densely}, we leverage the open source code and network parameter setting in \cite{geneva2020modeling}. The AR-DenseED model is constitutive of an encoding convolutional block, single dense block and a decoding block of different dense layers $\{4,3,4\}$, respectively. The growth rate is set as 4. We encode a 2D input variable consisting of five previous time-steps to latent features of half spatial dimensionality of input, and then latent features are decoded to the prediction of next time-step. In addition, we train the network for 1000 epochs using Adam with exponential decay rate of 0.995 after pre-training.

\subsubsection{Evaluation metrics}
To evaluate the solution accuracy produced by all the NN-based solvers, we assess the full-field error propagation in two phases: training and extrapolation. The definition of the full-field error $\epsilon_\tau$ at time $\tau$ is defined as the accumulative root-mean-square error (a-RMSE) given by
\begin{equation}
    \label{eq:error}
    \epsilon_\tau = \sqrt{\frac{1}{N_\tau} \sum_{k=1}^{N_\tau} \frac{\| \mathbf{u}^{*}(\mathbf{x},t_k) - \mathbf{u}^{\theta}(\mathbf{x},t_k) \|_2^2}{mn}},
\end{equation}
where $N_\tau$ is the total number of time steps within $[0,\tau]$, and $\mathbf{u}^{*}(\mathbf{x},t_k)$ is the reference solution.

\subsection{{2D Burgers' equations}}\label{s:burgers}
We firstly consider a 2D fluid dynamics problem, the famous Burgers' equations, given in the following tensor form: 
\begin{equation}
    \label{eq:burgers}
    \mathbf{u}_t + \mathbf{u} \cdot \nabla \mathbf{u} - \nu \Delta \mathbf{u} = \mathbf{0}, 
\end{equation}
where $\mathbf{u}=\{u,v\}$ denotes the fluid velocities; $\nu$ is the viscosity coefficient; $\Delta$ is the Laplacian operator. The 2D Burgers' equations describe the complex interaction of nonlinear convection and diffusion processes with possible shock behaviors, which usually acts as a benchmark model for comparing different computational algorithms. Herein, we choose $\nu=0.005$ and the spatial domain size as $\mathbf{x} \in [0,1]$.

\begin{figure}[t!]
    \centering
    \includegraphics[width=0.95\linewidth]{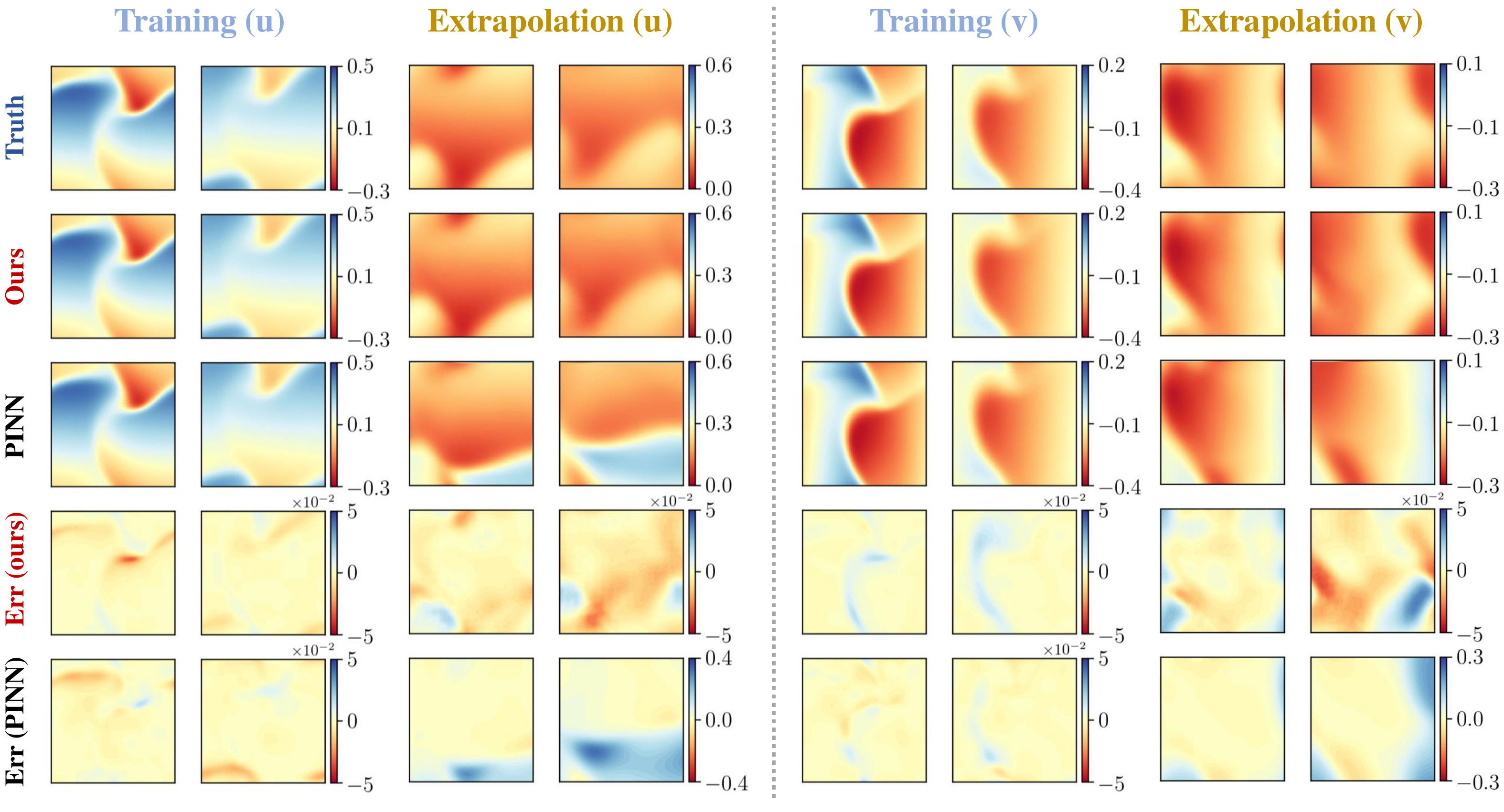}
    \caption{Comparison of solution accuracy and extrapolability between PhyCRNet and PINN for the 2D Burgers' equations. Four representative time instants are chosen for training ($t=1.0,2.0$) and extrapolation ($t=3.0,4.0$) phases. Err (ours) and Err (PINN) refer to the difference in the entire domain between ground truth reference and prediction by these two approaches.}
    \label{fig:comp_burgers}
\end{figure}

Moreover, the IC is generated from a Gaussian random field with periodic BCs using the open source code in \cite{li2020fourier}: $\mathbf{u}_0 \sim \mathcal{N}(0, 625(-\Delta+25\mathbf{I})^{-2})$. The ground truth reference solution is calculated using a high-order finite difference method with the 4th-order Runge-kutta time integration ($\delta t = 1\times 10^{-4}$). While for the PhyCRNet, we choose a relatively larger time interval $\delta t = 0.002$ over the discretized domain considering the implicit time marching in ConvLSTM. We train PhyCRNet to obtain the numerical solution of the Burgers' equations for 1000 time steps within the duration of $[0,2]$. The training time is 24 hours. Furthermore, based on the trained model, we predict the solution for another 1000 time steps (within time $[2,4]$) to test the extrapolability of our proposed method. The learning rate starts at $6\times10^{-4}$ and decays every 50 epochs by 1\%.

The solution snapshots predicted by PhyCRNet and PINN are shown in Figure \ref{fig:comp_burgers} in comparison with the ground truth reference. In particular, we select four representative snapshots in the training and extrapolation phases at $t=1.0,2.0$ and $t=3.0,4.0$ respectively for illustration. The solution error propagation with time for both methods is presented in Figure \ref{fig:err_burgers}. Firstly, it can be seen in Figure \ref{fig:comp_burgers} that the trained and extrapolated results using PhyCRNet both posses excellent agreement with the reference solution, while the PINN results fail to match the ground truth. In particular, we observe that most of the error field from PhyCRNet is close to zero, whereas the outcome of PINN exhibits much larger error especially on the boundaries (due to ``soft'' imposition of I/BCs). Secondly, the error propagation in Figure \ref{fig:err_burgers} further validates the superior solution accuracy of PhyCRNet with a-RMSE always below 0.01. PhyCRNet holds similar performance with PINN in training but leads in extrapolation, and prevails AR-DenseED both in training and extrapolation by two orders of magnitude. The result of AR-DenseED here turns out to be unsatisfactory on account of longer time evolution and larger spatial discretization compared to the 2D Burgers' case in \cite{geneva2020modeling}. Besides, the error propagation curve of PhyCRNet remains flattened along with time marching. The minor discrepancy between training and extrapolation shows the great potential of PhyCRNet for solution generalization.

\subsection{$\lambda$-$\omega$ RD equations}\label{s:lo}
The second example considered here is a $\lambda$-$\omega$ RD system in a 2D domain, which is widely known for its representation of multi-scale phenomenon for chemical and biological processes, e.g., turbulent behavior and self-organized patterns. Specifically, the two coupled nonlinear PDEs with the formation of spiral wave patterns are expressed as:
\begin{equation}
\begin{split}
    \label{eq:lo}
    u_t &= 0.1 \Delta u + \lambda(r)u - \omega(r)v, \\
    v_t &= 0.1 \Delta v + \omega(r)u + \lambda(r)v,
\end{split}
\end{equation}
where $u$ and $v$ are two field variables; $r=u^2+v^2$; $\lambda$ and $\omega$ are two real functions given by $\lambda=1-r^2$ and $\omega=-r^2$, respectively. The reference solution was generated using a spectral method \cite{rudy2017data} in the domain of $[-10,10]$ for 801 time steps ($\delta t=0.0125$). We train the model for 200 time steps with time duration of [0, 5], and perform the inference for [5, 10], where $\delta t=0.025$. The learning rate is set as $5\times10^{-4}$ and decays by 2\% every 100 epochs.

\begin{figure}[t!]
    \centering
    \includegraphics[width=0.95\linewidth]{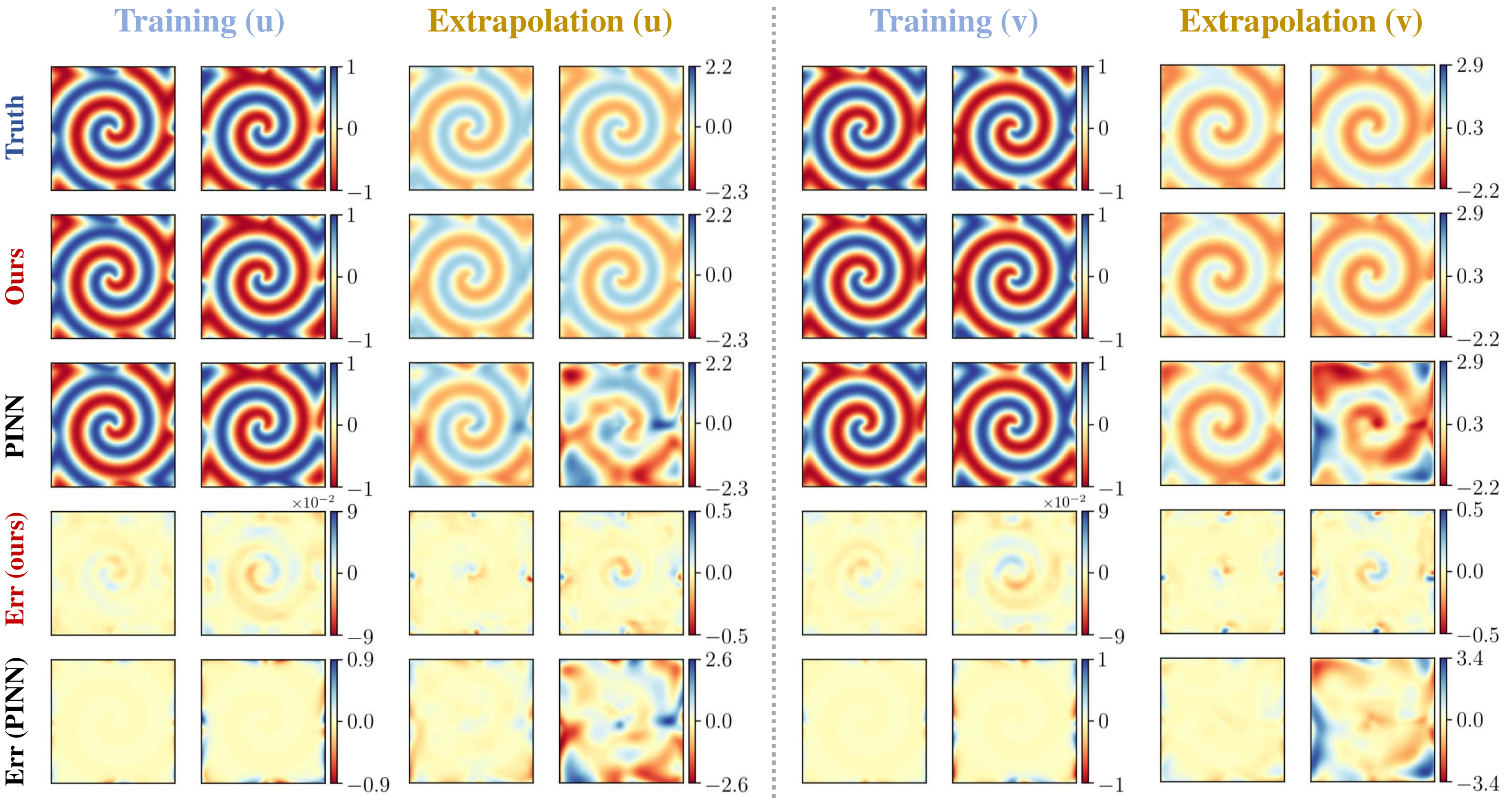}
    \caption{Comparison of solution accuracy and extrapolability between PhyCRNet and PINN for the $\lambda$-$\omega$ RD equations. Four representative time instants are chosen for training ($t=2.5,5.0$) and extrapolation ($t=7.5,10.0$) phases. Err(ours) and Err(PINN) refer to the difference in the entire domain between ground truth reference and these predicted by two DNN-based approaches.}
    \label{fig:comp_rd}
\end{figure}

The solution snapshots predicted by PhyCRNet and PINN, compared with the ground truth reference, are shown in Figure \ref{fig:comp_rd}. Given the simplicity of periodic patterns, both PhyCRNet and PINN produce good result in the training phase. We consider this phenomena coming from the truth that the portraits of this specific $\lambda$-$\omega$ RD system are simple and smooth, where the continuous approximation of solution by PINN plays a positive role. However, PINN yields large error on the boundaries and does not extrapolate well even for such simple patterns. On the contrary, PhyCRNet performs robustly with much smaller distributed errors even in the extrapolation phase. Moreover, Figure \ref{fig:err_rd} depicts the error propagation for PINN, AR-DenseED and PhyCRNet, where we observe that PhyCRNet outperforms both PINN and AR-DenseED with up to one order of magnitude smaller error.

\subsection{FitzHugh-Nagumo RD equations}\label{s:fn}
The final example aims to further test the capability of PhyCRNet for solving PDEs. Herein, we consider an explicable and mathematical portrait of neural excitation, namely, FitzHugh-Nagumo (FN) RD equations, written as: 
\begin{equation}
\begin{split}
    \label{eq:fn}
    u_t &= \gamma_u \Delta u + u - u^3 - v + \alpha, \\ 
    v_t &= \gamma_v \Delta v + \beta (u-v),  
\end{split}
\end{equation}
where $u$ and $v$ are two interactive components; $\gamma_u=1$ and $\gamma_v=100$ are diffusion coefficients, while $\alpha=0.01$ denotes the external stimulus and $\beta=0.25$ is the coefficient for reaction terms. With different diffusion and reaction coefficients, we can simulate varying neuron activities. The IC is generated by drawing random samples from a Gaussian distribution and the ground truth solution is produced by finite difference with the 4th-order Runge-Kutta time integration, in 2D domain $\Omega=[0,128]$ with $\delta t=2\times10^{-4}$. We attempt to train PhyCRNet to solve this PDE for 750 time steps with $\delta t=0.006$ (i.e., time duration $[0,4.5]$), and use the trained model to extrapolate the solution in $[4.5,9]$. The learning rate is set as $5\times10^{-5}$ and the decaying coefficient is 0.995 for every 50 epochs.

\begin{figure}[t!]
    \centering
    \includegraphics[width=0.95\linewidth]{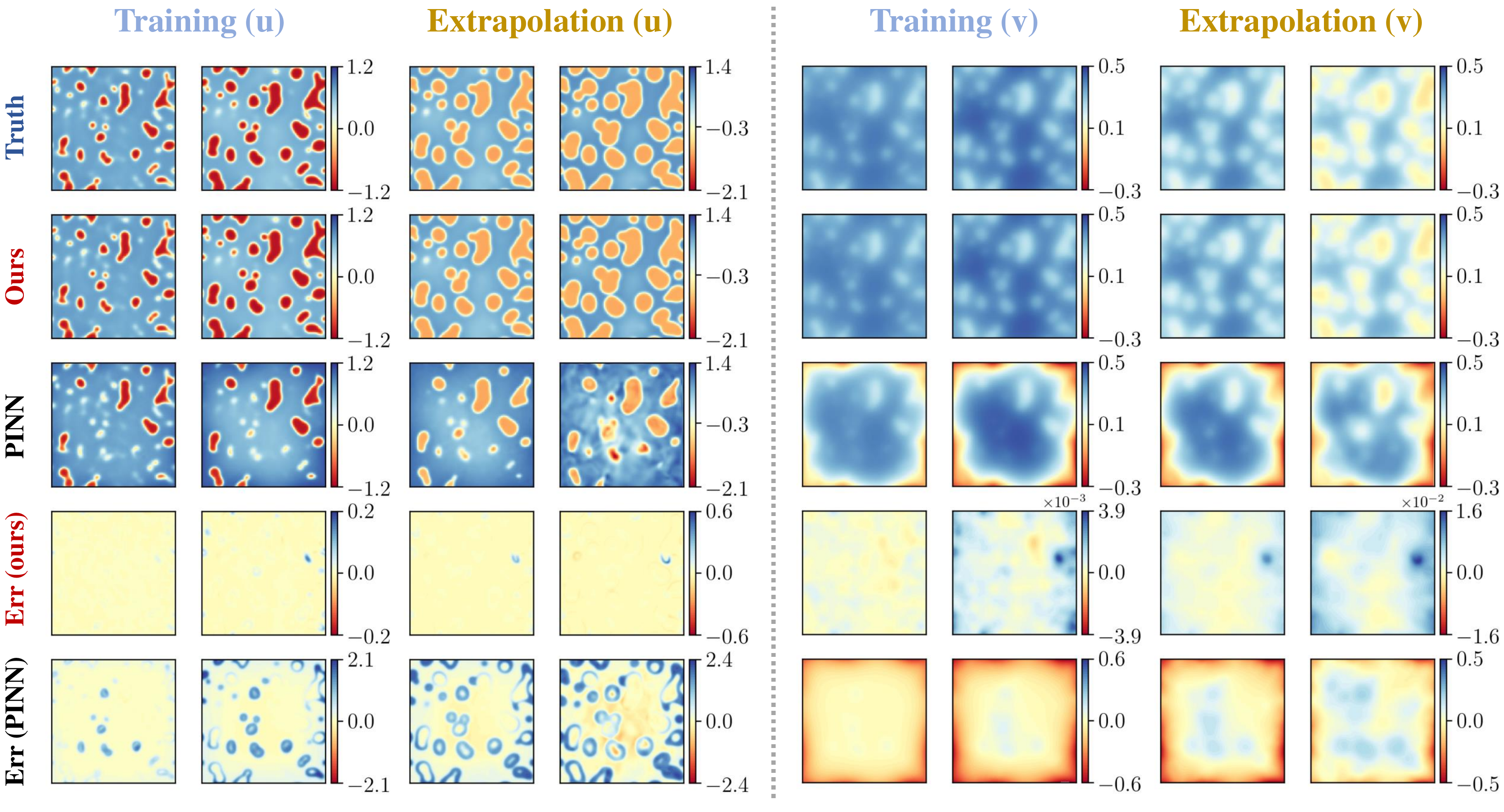}
    \caption{Comparison of solution accuracy and extrapolability between PhyCRNet and PINN for the FitzHugh-Nagumo RD Equations. Four representative time instants are chosen for training ($t=2.16,4.32$) and extrapolation ($t=6.48,8.64$) phases. Err(ours) and Err(PINN) refer to the difference in the entire domain between ground truth reference and these predicted by two DNN-based approaches.}
    \label{fig:comp_fn}
\end{figure}

\begin{figure}[t!]
	\centering
	    \subfigure[2D Burgers' equations]{\includegraphics[width=0.55\linewidth]{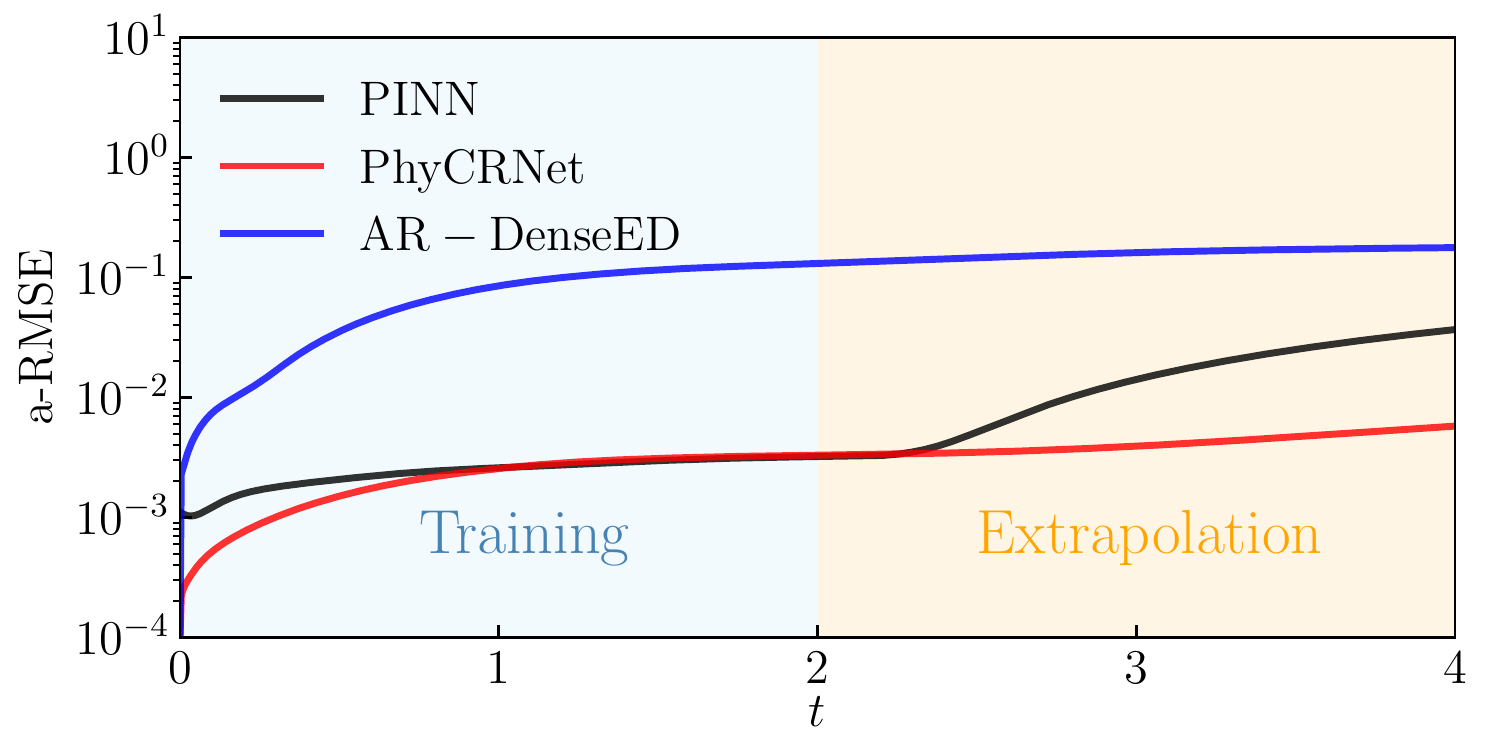}
	        \label{fig:err_burgers}} 
	    \hspace{0.5em}
	    \subfigure[$\lambda$-$\omega$ RD equations]{\includegraphics[width=0.55\linewidth]{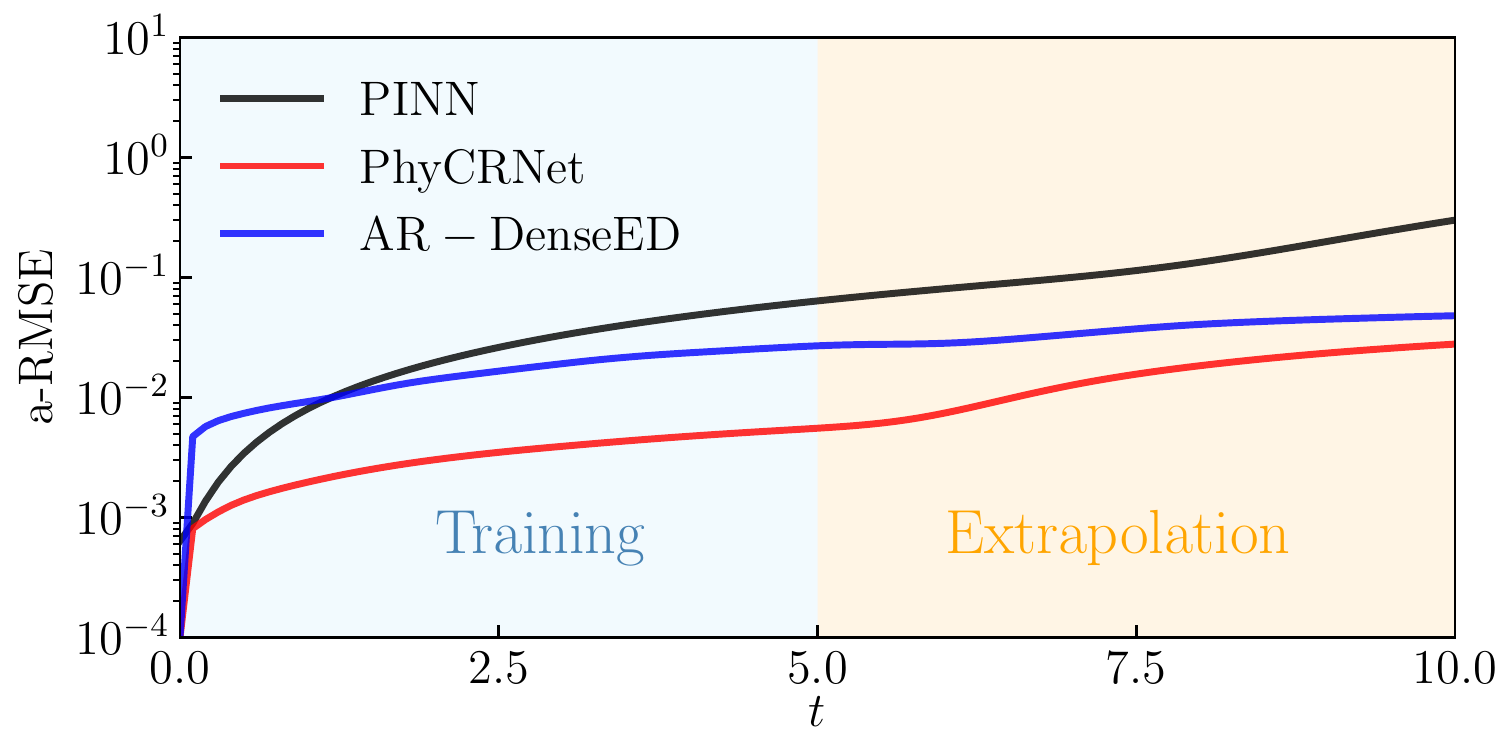}
	        \label{fig:err_rd}} 
	    \hspace{0.5em}
	    \subfigure[FN RD equations]{\includegraphics[width=0.55\linewidth]{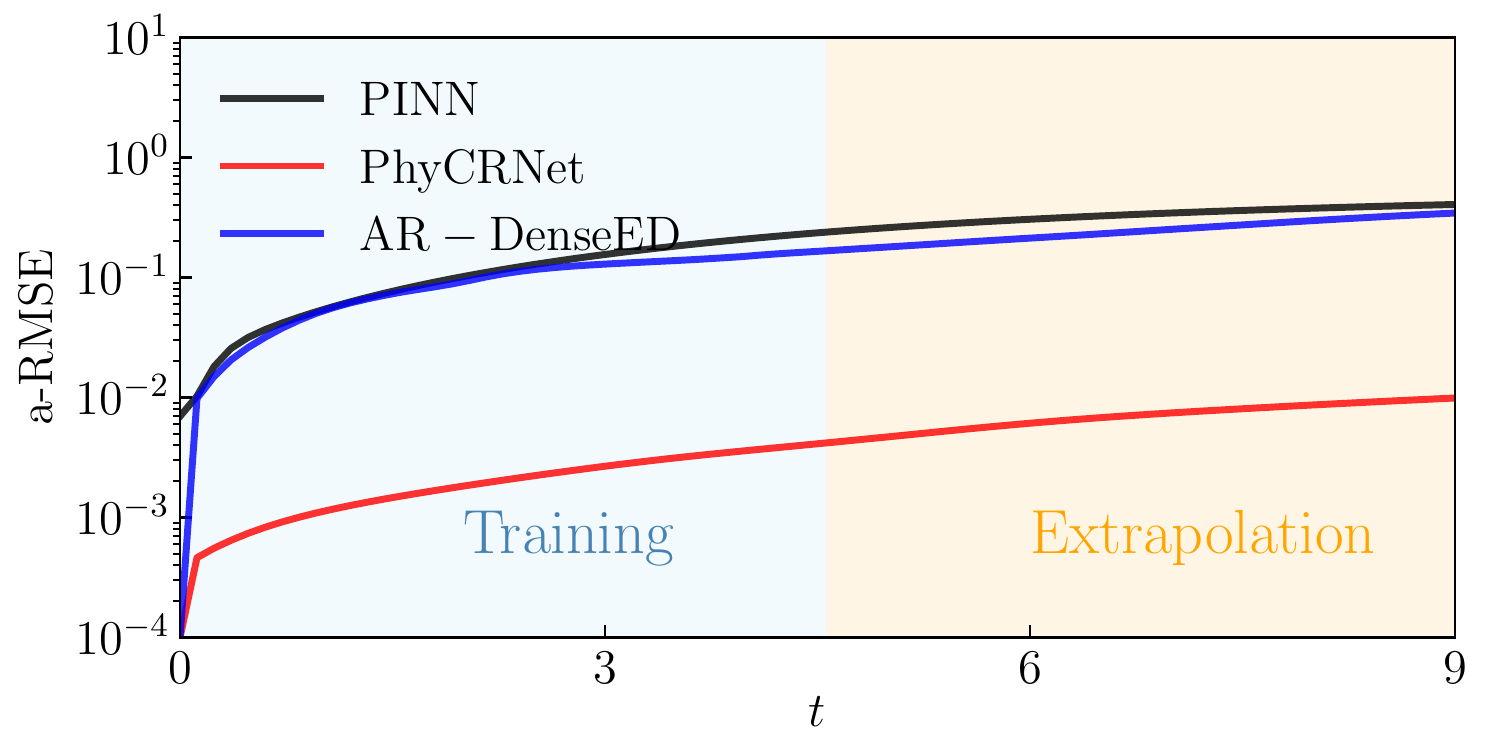}
	        \label{fig:err_fn}} 
	\vspace{0pt}
	\caption{Comparison of error propagation between PhyCRNet and PINN for three PDE systems. The curves in blue and orange areas present error propagation in the training and extrapolation phases, respectively.}
	\label{fig:err_prop}
\end{figure}

The solution snapshots predicted by PhyCRNet and PINN are shown in Figure \ref{fig:comp_fn}, along with the ground truth reference and error maps. It is notable that the FN system demonstrates more complex and interactive patterns compared with the previous two examples. Herein, PhyCRNet presents outstanding goodness of fit with the ground truth both in training and extrapolation, whereas PINN can barely simulate parts of the dynamical patterns. Moreover, the PhyCRNet error maps exhibit almost perfect results that are close to zeros, especially for the field variable $v$. The relatively large errors mainly cluster on propagating wave fronts. Nevertheless, considering the complex patterns of this PDE system, this kind of error is negligible. For the error propagation shown in Figure \ref{fig:err_fn}, PhyCRNet surpasses PINN and AR-DenseED up to two orders of magnitude. The a-RMSE of PhyCRNet increases mildly along with time but always keeps at a low level (i.e., $10^{-2}$), while the errors of PINN and AR-DenseED rise to the magnitude of $1$. Noteworthy, the error propagation in PhyCRNet extrapolation is relatively larger compared with the training phase, primarily due to the complicated interaction between the field components (e.g., with sharp propagating wave fronts). This problem can be alleviated by involving a longer time duration in training such that richer dynamics is captured. 

In general, PhyCRNet is highly capable of learning the spatiotemporal dependencies of PDE systems, ranging from fluid dynamics to RD systems. The salient outcomes of solution accuracy and extrapolation capability demonstrate PhyCRNet to be a novel and powerful method to solve complex multi-dimensional PDEs.

\subsection{Generalization to different ICs}\label{s:gen_ICs}
Theoretically, an ideal DNN-based solver should be capable of providing accurate solutions given any IC for a specific PDE. However, PINN fails to generalize to different ICs due to the intrinsic limitation of softly incorporating I/BCs into loss functions. Our proposed method (i.e., PhyCRNet/PhyCRNet-s) can generalize well outside the training phase, with respect to the system parameters and spatiotemporal discretization. This prominent strength comes from the architectural design of our networks, especially the AR input-output flow and hard-imposition of I/BCs. 

\begin{figure}[t!]
	\centering
	    \subfigure[IC $\#1$]{\includegraphics[width=0.23\linewidth]{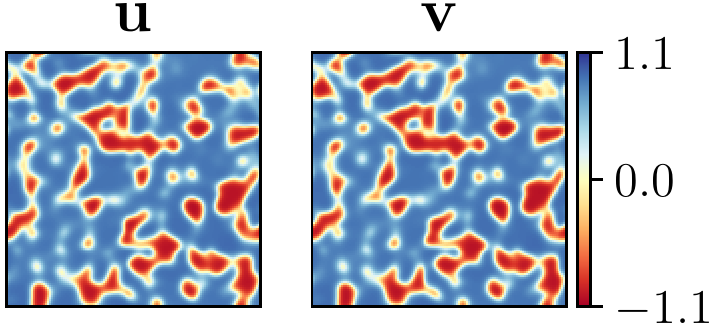}
	        \label{fig:ic1}} 
	    \hspace{0.05em}
	    \subfigure[IC $\#2$]{\includegraphics[width=0.23\linewidth]{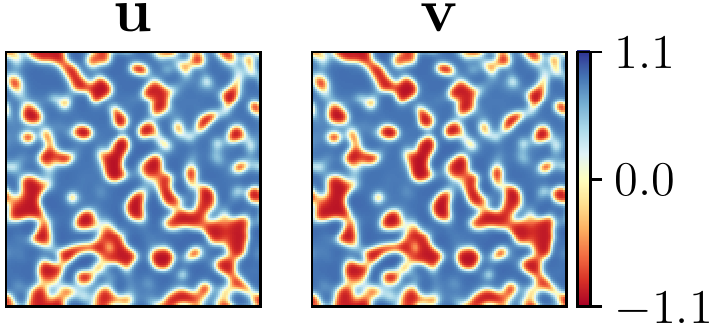}
	        \label{fig:ic2}} 
	    \hspace{0.05em}
	    \subfigure[IC $\#3$]{\includegraphics[width=0.23\linewidth]{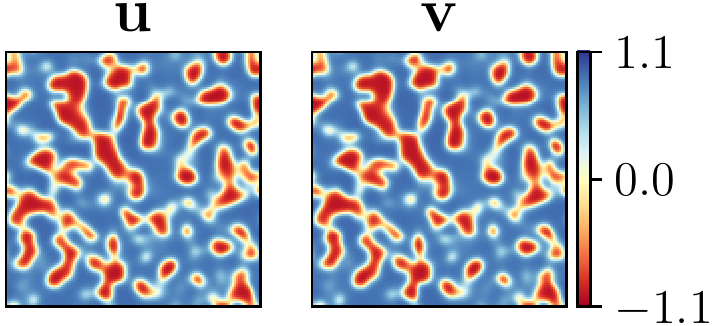}
	        \label{fig:ic3}} 
	    \hspace{0.05em}
	    \subfigure[IC $\#4$]{\includegraphics[width=0.23\linewidth]{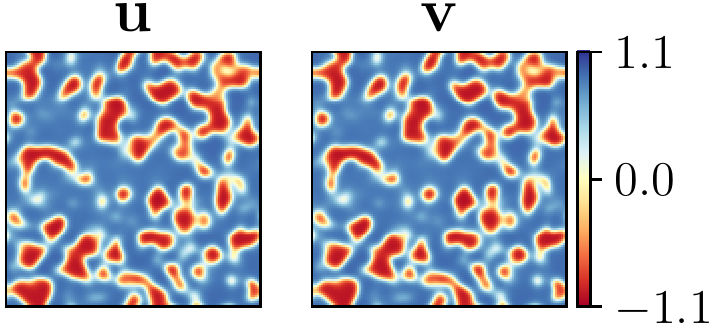} 
	        \label{fig:ic4}} 
	\vspace{0pt}
	\caption{Four randomly generated ICs.}
	\label{fig:diff_ics}
\end{figure}

\begin{figure}[t!]
    \centering
    \includegraphics[width=0.99\linewidth]{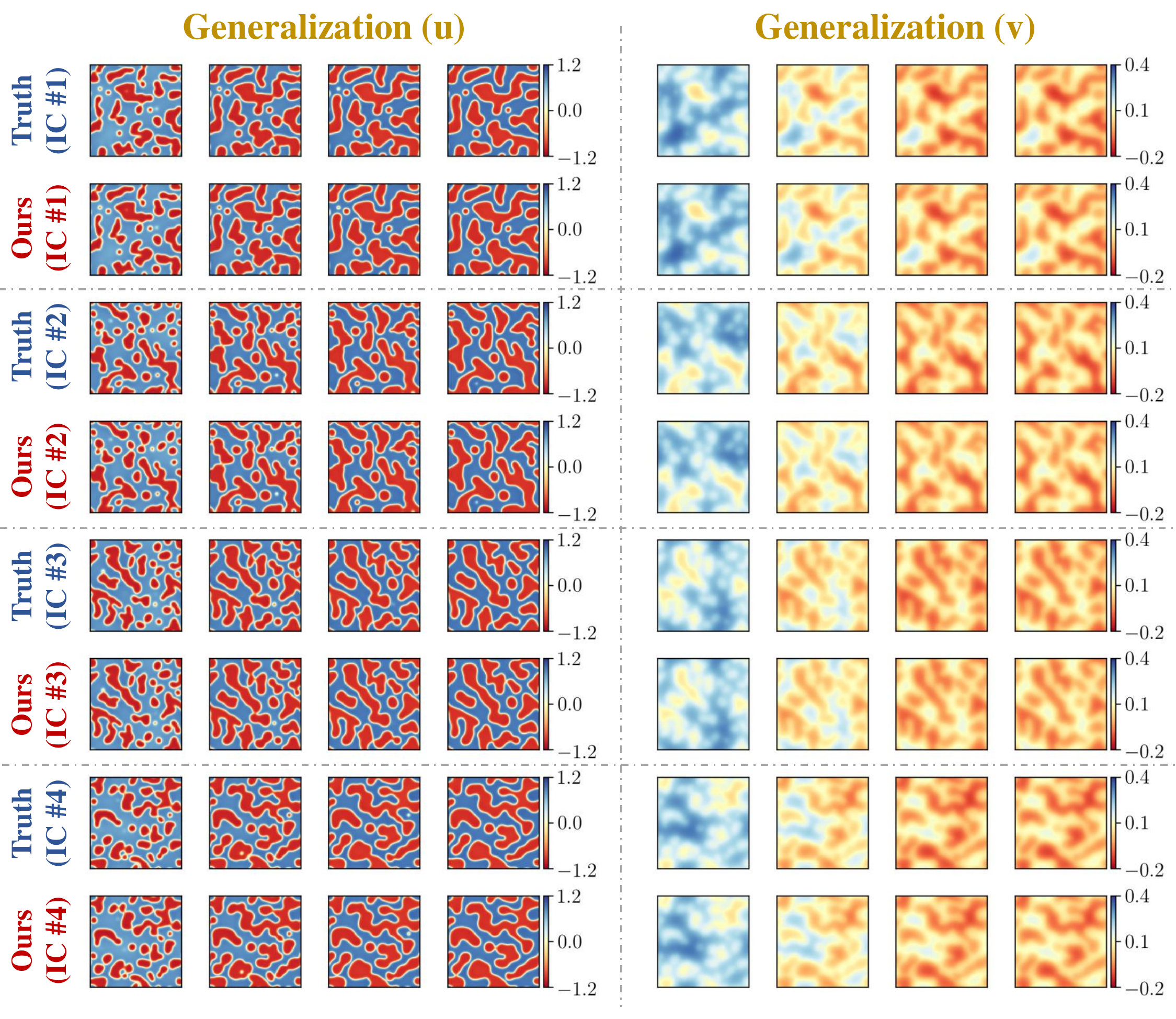}
    \caption{Generalization for four testing ICs of FN RD equations. Four representative time instants are chosen (i.e., $t=4.32,8.64,12.96,17.28$).}
    \label{fig:comp_fn_ics}
\end{figure}

\begin{figure}[t!]
	\centering
	    \subfigure[Generalization test]{\includegraphics[width=0.62\linewidth]{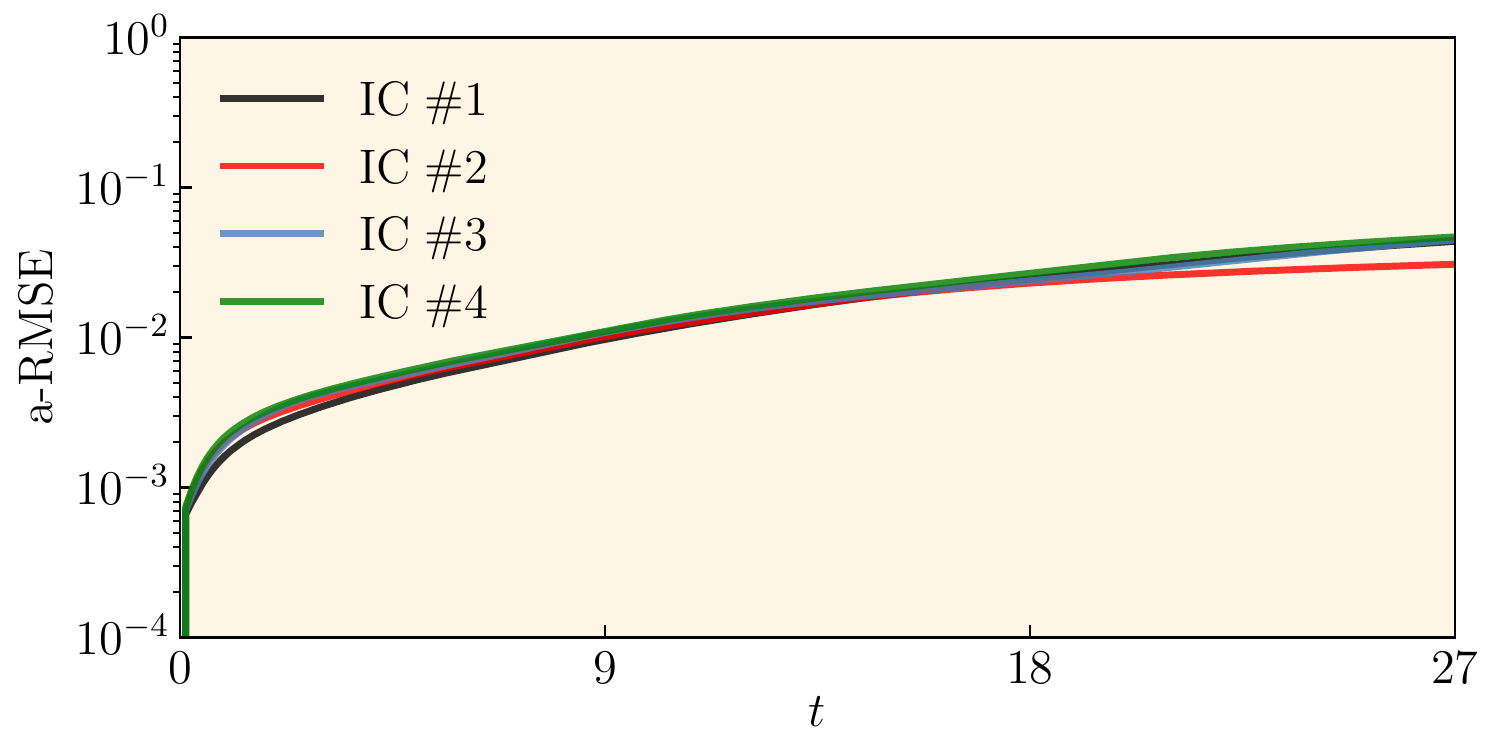}
	        \label{fig:err_fn_ics}} 
	    \hspace{0.5em}
	    \subfigure[Ablation study]{\includegraphics[width=0.62\linewidth]{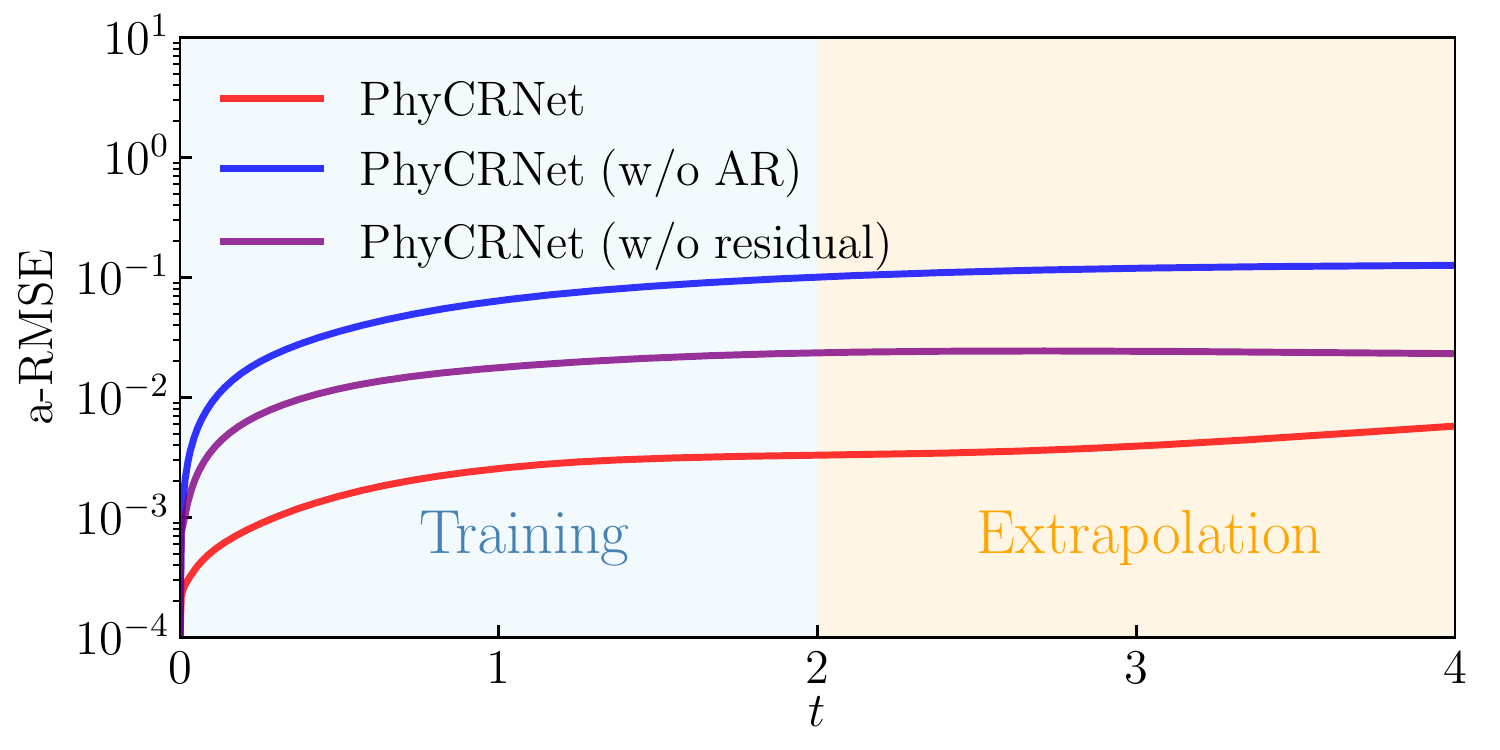}
	        \label{fig:ablation}} 
	    \hspace{0.5em}
	    \subfigure[Selection of $\mathcal{T}$]{\includegraphics[width=0.62\linewidth]{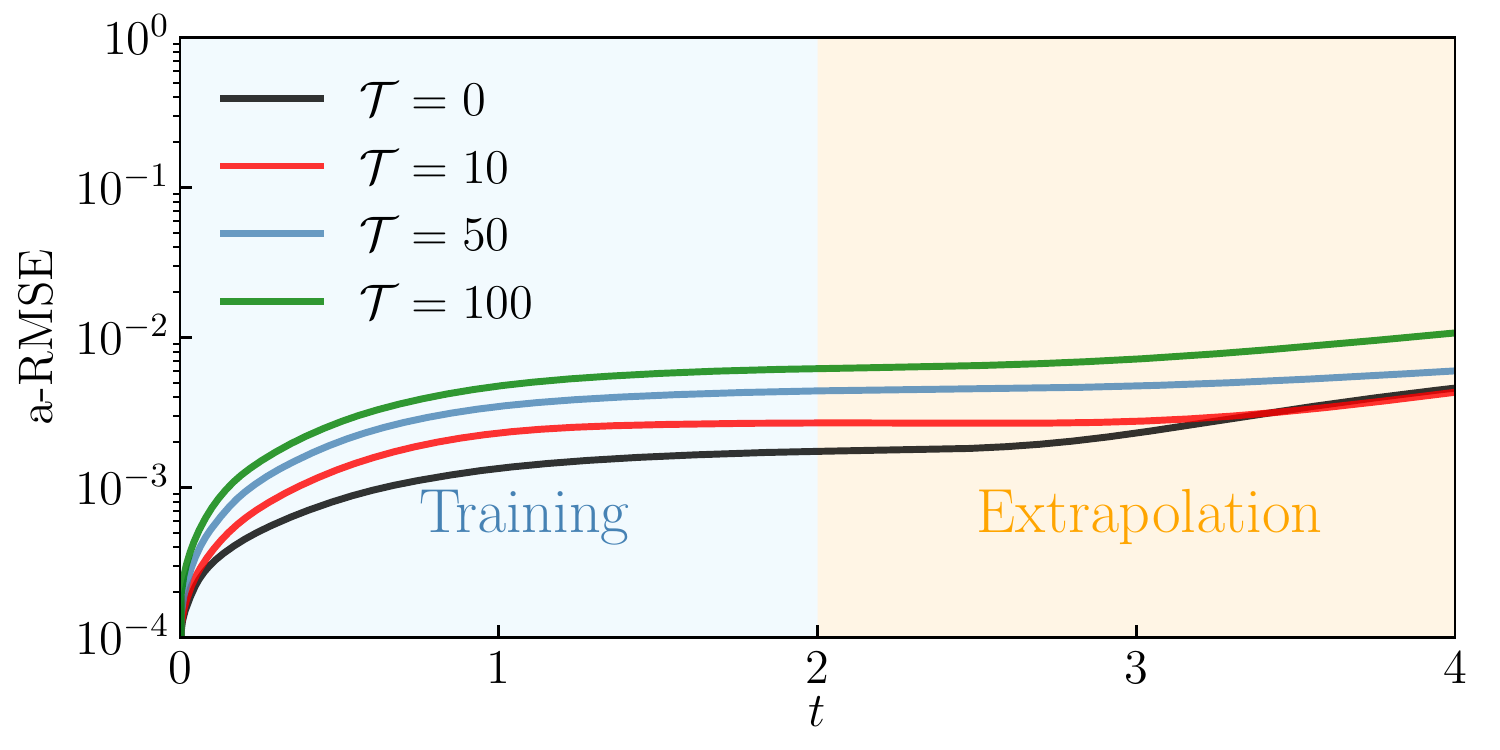}
	        \label{fig:comp_T}} 
	\vspace{0pt}
	\caption{(a) Error propagation for generalization testing at different ICs. $t$ is time duration ($[0,27]$). (b) Ablation study on neural architectures of PhyCRNet. (c) Error propagation of PhyCRNet (i.e., $\mathcal{T}=0$) and PhyCRNet-s (i.e., $\mathcal{T}=\{10,50,100\}$) frameworks.}
	\label{fig:err_prop_test}
\end{figure}

We evaluate the generalization performance of our model using the FN RD equations under four different ICs, as shown in Figure \ref{fig:diff_ics}. These four ICs are randomly sampled from Gaussian distribution with mean and standard deviation as 0 and 0.1, respectively. Compared with the extrapolation in Subsection \ref{s:fn}, we consider inferring a longer temporal evolution (i.e., 4500 time instants referring to the time duration [0, 27]) for all these IC scenarios. The testing results are presented in Figure \ref{fig:comp_fn_ics}, where we select four typical snapshots (i.e., $t=4.32,8.64,12.96,17.28$) of both $\mathbf{u}$ and $\mathbf{v}$ for illustration in comparison with the ground truth. All of the predicted results reveal the excellent abilities of our model to capture the evolutionary patterns and portray the local details. The positive evidence is also observed in error propagation described in Figure \ref{fig:err_fn_ics}. The error variances share similar tendency which increase smoothly and slowly, and are bounded below 0.04. Overall, these evaluation experiments suggest our model learns the underlying physical laws well and generalize to different ICs robustly.

\subsection{Ablation study} \label{s:ablation}
To justify the effectiveness of our designed frameworks, we implement an ablation study on PhyCRNet by solving the 2D Burgers' equations. Apart from the encoder-decoder, the rest of the network architecture consists of ConvLSTM, residual connection, input-output autoregressive (AR) scheme and filtering-based differentiation. Here we mainly investigate the contributions of AR scheme and global residual connection. The experimental setting includes three architectures: full PhyCRNet, PhyCRNet without AR scheme and PhyCRNet without global residual connection. We apply the unified training efforts for performance evaluation, which is exactly same in Section \ref{s:burgers}. The comparison result is described in Figure \ref{fig:ablation}. The structure of PhyCRNet without AR scheme works worst, obtaining larger errors by nearly two orders of magnitude compared with PhyCRNet. Additionally, PhyCRNet without the global residual connection also exhibits inferior performance both in the training and extrapolation stages, compared with the full model. Therefore, these two ablation experiments validate the significance and indispensability of AR scheme and global residual connection.

\subsection{Comparison between PhyCRNet and PhyCRNet-s}\label{s:net_comp}
Our idea of physics-informed convolutional-recurrent networks enjoys prominent flexibility in the architecture designing, where PhyCRNet and PhyCRNet-s are both proposed. More precisely, PhyCRNet is a special class of PhyCRNet-s with $\mathcal{T}=0$. The skip-encoder strategy with respect to different $\mathcal{T}$ can effectively provide a balance/tradeoff between accuracy and efficiency. 

In this part, we investigate the effects of hyper-parameter $\mathcal{T}$ to the performance of network architectures, considering the 2D Burgers' equations as a testing example. Herein, we select four different architectural structures with $\mathcal{T}=\{0,10,50,100\}$ respectively, and keep other neural components fixed. All of the four experiments here undergo the same training procedures, which include 10,000 epochs using Adam after pre-training with the learning rate starting at $6\times10^{-4}$ and decaying every 50 epochs by 1\%. 

The evaluation results are presented in Table \ref{tab:comp_phycrnet} and Figure \ref{fig:comp_T}. Herein, we define a relative full-field error with respect to the ground truth, in order to assess the model performance in training and extrapolating stages. As shown in Table \ref{tab:comp_phycrnet}, the general trend is that with the parameter $\mathcal{T}$ increasing, the training effort (i.e., computational time) decreases mildly and the solution errors both in training and extrapolation grow. However, there is also one surprising observation that PhyCRNet works inferiorly than PhyCRNet-s ($\mathcal{T}=10$) in the extrapolation, though leads in training solution accuracy. We think the reason behind this interesting phenomenon is that a small $\mathcal{T}$ helps relax the model training, which enhances the extrapolation ability. Moreover, the details of error propagation described in Figure \ref{fig:comp_T} further validates the significant result. In the training phase, the model with smaller $\mathcal{T}$ behaves more excellent. Nevertheless, when extrapolating relatively more time steps, PhyCRNet becomes weaker against ($\mathcal{T}=10$). Therefore, we conclude that a reasonably small $\mathcal{T}$ contributes to the balance between accuracy and efficiency as well as the robustness for extrapolation.

\begin{table}
  \caption{Performance comparison between PhyCRNet and PhyCRNet-s.}
  \label{tab:comp_phycrnet}
  \centering
  {\small
  \begin{tabular}{c c c c}
    \toprule
    Model  & Time [s]/epoch & Training error [\%] & Extrapolating error [\%]\\
    \midrule
    PhyCRNet  & 8.64 & 0.83 & 3.40\\
    \midrule
    PhyCRNet-s ($\mathcal{T}=10$)  & 8.18 & 1.28 & 2.97\\
    \midrule
    PhyCRNet-s ($\mathcal{T}=50$)  & 8.04 & 2.09 & 3.93\\
    \midrule
    PhyCRNet-s ($\mathcal{T}=100$) & 7.94 & 2.94 & 7.52\\
    \bottomrule
  \end{tabular}}
\end{table}

\section{Discussion}\label{s:discussion}
In this section, we aim to make a comprehensive and comparative discussion between our discrete methods and the standard continuous approaches (i.e., PINNs), as well as provide future perspectives based on the current investigation. Prior studies \cite{raissi2019physics,raissi2020hidden} have shown the great potential of PINNs for forward and inverse analysis of spatiotemporal PDEs, but the comparison against discrete methods are still lacking. Generally, there are two distinctive 
aspects, mainly covering mesh requirement and encoding of I/BCs. 

First of all, as a continuous learning method, PINNs hold the apparent strength of meshfree, while our PhyCRNet and PhyCRNet-s frameworks rely on prescribed spatiotemporal mesh and are limited to regular grids. However, thanks to the emerging geometric learning methods (e.g., graph neural networks \cite{kipf2016semi,bresson2017residual,bronstein2017geometric,battaglia2018relational,defferrard2016convolutional}), we can extend the current architectures to irregular meshes/grids, making them more versatile. Besides, due to the way of functional approximation with global basis, PINNs are dedicated for ``global learning'' which helps capture the general dynamical patterns of nonlinear PDE systems. Discrete learning methods focus on local features within the physical domains because of the employment of convolutional kernels, which provides a more reliable way of capturing local morphology in the PDE solution.

Secondly, the enforcement of I/BCs significantly affects the solution accuracy of PDEs, especially on the boundaries. For PINNs, the soft imposition of I/BCs, which is presented as a penalty factor in loss function, requires massive hyper-parameter tuning for the optimal weight of I/BC loss component. Usually, such a soft strategy cannot guarantee the solution accuracy on the boundaries, and fail to generalize to different ICs as discussed in Section \ref{s:experiment}. On the other hand, for our discrete approaches, it is easy and suitable to rigorously incorporate the discrete boundary values into the network via an intrinsic padding scheme, which painlessly and remarkably promotes the solution accuracy on the boundaries. In addition, instead of integrating IC loss into the loss function, we define the IC as the first input state variable for the network, which makes generalization to different IC scenarios achievable. Apart from the I/BCs, the known conservation laws (e.g., mass conservation) can also be enforced simultaneously if applicable. For example, strict enforcement of mass conservation can be realized by applying a stream function as the solution variable in the network for fluid dynamics. Besides, it is more significant and realistic to design a BC encoding strategy for handling various BCs. The idea is to utilize parametric learning, where parametric BCs are encoded to serve as an additional input to the network.

\section{Conclusion}\label{s:conclusion}
Solving PDEs is fundamental to a wide range of scientific computational problems, where physics-informed DNNs show promise to tackle relevant challenges especially in inverse problems and data assimilation. The majority of existing methods may suffer from issues of scalability, error propagation and generalization. This motivates us to develop a novel PhyCRNet/PhyCRNet-s architecture as a universal model for solving spatiotemporal PDEs. The hard enforcement of I/BCs via designated padding prompts a well-posed optimization problem in network training, improves the solution accuracy and facilitates convergence. Finally, we validate the excellent performance of our proposed methods in solution accuracy, extrapolability, and generalizability. These proposed networks have promising potential to serve as a reliable surrogate model for data assimilation and inverse analysis of physical systems where data is scarce and noisy, which will be demonstrated in the next step of our study. Furthermore, our current networks can be naturally extended to tackle irregular spatial domains by incorporating graph neural networks, as well as modified from forward Euler scheme to high-order difference strategy (e.g., high-order Runge-Kutta scheme) for more accurate temporal evolution modeling.

\section*{Acknowledgement}
Pu Ren would like to acknowledge the support by the Vilas Mujumdar Fellowship at Northeastern University. Yang Liu acknowledges the support by the TIER 1 Seed Grant Program at Northeastern University. Hao Sun acknowledges the support by the Engineering for Civil Infrastructure program at National Science Foundation under grant CMMI-2013067 and the research award from MathWorks.

\section*{Data Availability}
The datasets and computer codes are available on \url{https://github.com/isds-neu/PhyCRNet} upon final publication of the paper.

\bibliographystyle{elsarticle-num}
\bibliography{refs.bib}

\end{document}